\documentclass[journal, twoside]{IEEEtran}
\usepackage[pdftex]{graphicx}
\usepackage[numbers]{natbib}

\usepackage{booktabs}
\usepackage{moreverb}
\usepackage{url}
\usepackage{amsmath}
\usepackage{bm}
\usepackage{subfig}

\usepackage[colorlinks,bookmarksopen,bookmarksnumbered,citecolor=red,urlcolor=red]{hyperref}
\hypersetup
{
	pdftitle = {Motion planning for quadrupedal locomotion: coupled planning, terrain mapping and whole-body control},
	pdfauthor = {Carlos Mastalli},
	pdfsubject = {T-RO paper},
	pdfkeywords = {trajectory optimization, legged robots, challenging locomotion, whole-body control and terrain mapping},
	pdftoolbar = true,
	colorlinks = true,
	linkcolor = black,
	citecolor = black,
	urlcolor = black,
}
\newcommand{\newtext}[1]{\textcolor{black}{#1}}
\newcommand{\revisedtext}[1]{\textcolor{black}{#1}}
\usepackage{color}
\usepackage{amsfonts}
\usepackage{amssymb}
\usepackage[usenames,dvipsnames]{xcolor}%\usepackage{xcolor,colortbl}
\definecolor{blue_iit}{RGB}{0,0,0}%{RGB}{51,51,255}
\usepackage{algpseudocode}
\usepackage{algorithm}
\usepackage{glossaries}
	\newacronym{hyq}{HyQ}{Hydraulically actuated Quadruped}
	\newacronym{lf}{LF}{Left-Front}
	\newacronym{rf}{RF}{Right-Front}
	\newacronym{lh}{LH}{Left-Hind}
	\newacronym{rh}{RH}{Right-Hind}
	\newacronym{ptu}{PTU}{Pan and Tilt Unit}
	\newacronym{imu}{IMU}{Inertial Measurement Unit}
	\newacronym{dofs}{DoFs}{Degree of Freedoms}
	\newacronym{rt}{RT}{Real Time}
	\newacronym{com}{CoM}{Center of Mass}
	\newacronym{cop}{CoP}{Center of Pressure}
	\newacronym{zmp}{ZMP}{Zero Moment Point}
	\newacronym{icp}{ICP}{Instantaneous Capture Point}
	\newacronym{cmp}{CMP}{Centroidal Moment Pivot}
	\newacronym{grfs}{GRFs}{Ground Reaction Forces}
	\newacronym{mcot}{\revisedtext{ELC}}{\revisedtext{Estimated Locomotion Cost}}
	\newacronym{cmm}{CMM}{Centroidal Momentum Matrix}
	\newacronym{rnea}{RNEA}{Recursive Newton-Euler Algorithm}
	\newacronym{slip}{SLIP}{Spring Loaded Inverted Pendulum}
	\newacronym{eom}{EoM}{Equation of Motions}
	\newacronym{aras}{ARA$^*$}{Anytime Repairing A$^*$}
	\newacronym{mpc}{MPC}{Model Predictive Control}
	\newacronym{qp}{QP}{Quadratic Programming}
	\newacronym{sqp}{SQP}{Sequential Quadratic Programming}
	\newacronym{micp}{MICP}{Mixed-Integer Convex Programming}
	\newacronym{cmaes}{CMA-ES}{Covariance Matrix Adaptation Evolution Strategy}
	\newacronym{ara}{ARA*}{Anytime Repairing A*}
	\newacronym{pca}{PCA}{Principal Component Analysis}
	\newacronym{cpg}{CPG}{Central Pattern Generator}
	\newacronym{wbc}{WBC}{Whole-Body Control}
\usepackage[tight]{units}
\newcommand{\sref}[1]{Section~\ref{#1}}
\newcommand{\fref}[1]{Fig.~\ref{#1}}
\newcommand{\eref}[1]{Eq.~(\ref{#1})}
\newcommand{\tref}[1]{Table~\ref{#1}}
\usepackage{cleveref}
\crefname{figure}{Fig.}{Fig.}
\crefname{equation}{Eq.}{Eq.}
\AtBeginDocument{%
}
\newcommand{\Rnum}{\mathbb{R}} % Symbol fo the real numbers set

\newcommand{\vc}[1]{\mathbf{\bm{#1}}} 					% Vector symbol
\DeclareMathOperator*{\argmin}{\arg\!\min}				% argmin
\newcommand{\mat}[1]{\ensuremath{\begin{bmatrix}#1\end{bmatrix}}}	% matrix
\newcommand\BibTeX{{\rmfamily B\kern-.05em \textsc{i\kern-.025em b}\kern-.08em
T\kern-.1667em\lower.7ex\hbox{E}\kern-.125emX}}

\title{Motion Planning for Quadrupedal Locomotion: Coupled Planning, Terrain Mapping and Whole-Body Control}
\author{Carlos Mastalli\quad Ioannis Havoutis\quad Michele Focchi\quad Darwin G. Caldwell\quad Claudio Semini
\thanks{Manuscript received March 10, 2020; Accepted June 8, 2020.
This work was mainly supported by the Istituto Italiano di Tecnologia, as well as by MEMMO and HyQ-REAL European projects, and The Alan Turing Institute. MEMMO is a collaborative project supported by European Union within the H2020 Program, under Grant Agreement No. 780684. HyQ-REAL was part of the EU’s Seventh Framework Programme for research, technological development and demonstration under grant agreement No. 601116 which belongs to the ECHORD++ (The European Coordination Hub for Open Robotics Development).
This article was recommended for publication by Associate Editor P.-C. Lin and Editor E. Yoshida upon evaluation of the reviewers’ comments.
\textit{(Corresponding author: Carlos Mastalli.)}
}
\thanks{
Carlos Mastalli is with the Dynamic Legged Systems (Lab), Istituto Italiano di Tecnologia, 16163 Genova, Italy, and also with the School of Informatics, University of Edinburgh, South Bridge EH8 9YL, U.K. (e-mail: \href{mailto:carlos.mastalli@ed.ac.uk}{carlos.mastalli@ed.ac.uk}).
}
\thanks{
Ioannis Havoutis is with the Oxford Robotics Institute, Department of Engineering Science, University of Oxford, Oxford OX1 2JD, U.K. (e-mail: \href{mailto:ioannis@robots.ox.ac.uk}{ioannis@robots.ox.ac.uk}).
}
\thanks{
Michele Focchi and Claudio Semini are with the Dynamic Legged Systems (Lab), Istituto Italiano di Tecnologia, 16163 Genova, Italy (e-mail: \href{mailto:michele.focchi@iit.it}{michele.focchi@iit.it}; \href{mailto:claudio.semini@iit.it}{claudio.semini@iit.it}).
}
\thanks{
Darwin G. Caldwell is with the Department of Advanced Robotics, Istituto Italiano di Tecnologia, 16163 Genova, Italy (e-mail: \href{mailto:darwin.caldwell@iit.it}{darwin.caldwell@iit.it}).
}
\thanks{
The manuscript contains simulation and experimental trials and simulations which helps the readers to easy understand the motion planning and control methods. Contact \href{mailto:carlos.mastalli@ed.ac.uk}{carlos.mastalli@ed.ac.uk} for further questions about this work.
}}
\begin{document}

\maketitle

%------------------------------ ABSTRACT 
\begin{abstract}%150-250 word abstract
\revisedtext{Planning whole-body motions while taking into account the terrain conditions} \newtext{is a challenging problem for legged robots} since \revisedtext{the terrain model might produce many local minima}.
Our coupled planning method \revisedtext{uses stochastic and derivatives-free search} to \revisedtext{plan} both foothold locations and \revisedtext{horizontal} motions \revisedtext{due to the local minima produced by the terrain model}.
It jointly optimizes body motion, step duration and foothold selection, and \revisedtext{it models the terrain as a cost-map}.
\revisedtext{Due to the novel attitude planning method, the horizontal motion plans} can be \revisedtext{applied} to various terrain conditions.
\revisedtext{The attitude planner ensures the robot stability by imposing limits to the angular acceleration}.
Our whole-body controller tracks compliantly trunk motions while avoiding slippage, as well as kinematic and torque limits.
\revisedtext{Despite the use of a simplified model, which is restricted to flat terrain, our approach shows remarkable capability to deal with a wide range of non-coplanar terrains.}
\revisedtext{The results are validated by} experimental \revisedtext{trials} and comparative evaluations \revisedtext{in a series} of terrains of progressively increasing \revisedtext{complexity}.
\end{abstract}

\begin{IEEEkeywords}
\revisedtext{legged locomotion, trajectory optimization, challenging terrain, whole-body control and terrain mapping}
\end{IEEEkeywords}

%------------------------------ INTRODUCTION
\section{Introduction}\label{sec:introduction}
\IEEEPARstart{L}{egged} robots can deliver substantial advantages in real-world environments \newtext{by offering} mobility that is unmatched by wheeled counterparts.
Nonetheless, most legged robots are still confined to structured terrain.
\revisedtext{One of the main reasons} is the difficulty on generating complex dynamic motions \revisedtext{while considering} the terrain conditions.
\revisedtext{Due to this complexity}, many legged locomotion approaches focus on \newtext{terrain-blind methods} \revisedtext{with instantaneous actions}~\citep[e.g.][]{Wensing2013,Barasuol2013,Bellicoso2017}.
\newtext{These \revisedtext{heuristic approaches} assume that reactive actions are enough to ensure the robot stability under unperceived terrain conditions.}
\revisedtext{Unfortunately,} these approaches \newtext{cannot} tackle all \newtext{types of} terrain, \revisedtext{in particular terrains with big discontinuities}.
Such difficulties have \revisedtext{limited} the use of legged systems to \newtext{specific terrain topologies}.

Trajectory optimization with contacts \revisedtext{has} gained attention in the legged robotics community~\citep{Dai2016,Aceituno2017b,Winkler2018}.
\revisedtext{It} aims to overcome the drawbacks of \newtext{terrain-blind} approaches by considering a horizon of future events (e.g. body movements and foothold locations).
\revisedtext{It} could potentially improve the robot stability given a certain terrain.
\revisedtext{However,} in spite of the promising benefits, most of the works are focused on flat conditions or on simulation.
\revisedtext{For instance, these trajectory optimization methods do not incorporate any \textit{terrain-risk} model.}
\revisedtext{This model} serves to quantify the footstep difficulty and uncertainty.
\revisedtext{Nonetheless}, \newtext{it is not \revisedtext{yet} clear how to properly incorporate \revisedtext{this model} inside a trajectory optimization framework.
Reason why terrain models are often used only for foothold planning (decoupled approach)~\citep[e.g.][]{Kolter2008b,Kalakrishnan2010b}.}

\subsection{Contribution}
\revisedtext{To address challenging terrain locomotion,}
\revisedtext{we extend our previous planning method~\citep{Mastalli2017} in two ways.}
\revisedtext{First,} we propose a novel robot attitude planning method that heuristically \revisedtext{adapts} trunk
orientation while still guaranteeing the robot's stability.
\revisedtext{Our approach establishes limits in the angular acceleration that keep the estimated \gls{cmp} inside the support region.}
With our attitude planner, the robot can cross \revisedtext{challenging} terrain with height elevation changes.
\revisedtext{It allows the robot to navigate over stairs and ramps, as shown in the experimental and simulation trials.}
\revisedtext{Second}, we propose \revisedtext{a} terrain model (based on log-barrier functions) that \revisedtext{robustly describes} feasible \revisedtext{footstep locations}.
\revisedtext{This work presents first experimental studies on how both models influence} the legged locomotion over challenging terrain.
\revisedtext{The paper presents an exhaustive comparison of the coupled planning described in this work against a decoupled planning method proposed in~\citep{Winkler2015,Mastalli2015}}.
For doing so, we integrate online terrain mapping, state estimation and whole-body control.
This article is an extension of \newtext{earlier results}~\citep{Mastalli2017} presented at the IEEE International Conference on Robotics and Automation (ICRA) 2017.

The remainder of the paper is structured as follows: after discussing previous research in the field of dynamic whole-body locomotion (\sref{sec:related_work}) we briefly describe our decoupled planner method, which we use for comparison. 
Next, we \revisedtext{introduce our locomotion framework in~\sref{sec:system}.}
We describe our coupled planning method (\sref{sec:coupled_planning}) \newtext{and how the terrain model is formulated in our trajectory optimization}.
\sref{sec:execution} \newtext{briefly describes} a controller designed for dynamic motions.
\revisedtext{This controller improves the tracking performance and the robustness of the locomotion by passivity-based control paradigm.}
In \sref{sec:results},~\ref{sec:discussion} we evaluate the performance of our \revisedtext{locomotion framework}, and \revisedtext{provide comparison with our decoupled planner}, in real-world experimental trials and simulations.
Finally, \sref{sec:conclusion} summarizes this work.

%------------------------------ RELATED WORK 
\section{Related work}\label{sec:related_work}
In environments where smooth \revisedtext{and} continuous support is available (\revisedtext{floors}, fields, roads, etc.), exact foot placement is not crucial \revisedtext{in} the locomotion process.
\revisedtext{Typically,} legged \revisedtext{robots are free to move with a \textit{gaited} strategy, which only considers the balancing problem}.
The early work of Marc Raibert~\citep{Raibert1986} crystallized \revisedtext{these} principles of dynamic locomotion and balancing.
\revisedtext{Going beyond the flat terrain,} the \emph{\newtext{Spot}} and \emph{\newtext{SpotMini}} quadrupeds are a recent extension of this work.
While \emph{\newtext{SpotMini}} is able to traverse irregular terrain using a reactive controller, \newtext{we believe that} \revisedtext{(as there is no published work)} the footholds are not planned in advance.
Similar performance can be seen on the \gls{hyq} robot, that is able to overcome obstacles with reactive controllers~\citep{Barasuol2013,Havoutis2013a} \revisedtext{and/}or step reflexes~\citep{Focchi2013,Focchi2017a}.
% \revisedtext{We could also list other interesting works in the literature, for example, in galloping~\citep{Park2015} and bounding~\citep{Park2014,Orsolino2017}.}

\revisedtext{The main limitation of those gaited approaches is that they quickly reach the robot limits (e.g. torque limits) in environments with complex geometry: large gaps, stairs or rubble, etc.}
\revisedtext{Furthermore, in these environments, the robot often can} afford only few possible discrete footholds.
\revisedtext{Reason why it is important to carefully select footholds that do not impose a particular gaited strategy.}
\revisedtext{Towards this direction,} the DARPA Learning Locomotion Challenge stimulated the development of \revisedtext{strategies that handle a variety of} terrain conditions.
It resulted in a number of successful control architectures~\citep{Kolter2008b,Kalakrishnan2010b} \revisedtext{that} plan~\citep{Pongas2007,Zucker2011,Shkolnik2011} and execute footsteps~\citep{Rebula2007} \revisedtext{in a predefined set of} challenging terrains.
\newtext{\revisedtext{Roughly speaking,} these approaches \revisedtext{are able} to compute foothold locations by using tree-search algorithms, and \revisedtext{to} learn the terrain \revisedtext{cost-map} from user demonstrations~\cite{Kalakrishnan2010a}.}

\newtext{Legged locomotion can \revisedtext{also} be formulated as \revisedtext{an} optimal control problem.
\revisedtext{However, most works do not consider the contact location and timings~\citep[e.g.][]{Carpentier2017,Ponton2016,Budhiraja2018} due to the requirement of having a smooth formulation.}
The contact location and timings are often planned using heuristic rules with partial guarantee of dynamic feasibility~\cite{Tonneau2018,Deits2014}.
\revisedtext{Using these rules, it is possible to avoid the combinatorial complexity and the excessive computation time of more formal approaches (\citep[e.g.][]{Tassa2010,Mordatch2012a,Posa2013,Dai2014}).}
\revisedtext{Even though} recent works have reduced the computation time by a few orders of magnitude~\citep[e.g.][]{Aceituno2017b,Winkler2018}, they are still limited to offline planning and they require a convex model of the terrain.
\revisedtext{In the following subsection}, we briefly describe our previous decoupled planning method, which \revisedtext{will be} used \revisedtext{as baseline} to compare \revisedtext{against} our \revisedtext{new} coupled planning method.}

\subsection{Decoupled planning}\label{sec:decoupled_planning}
% \begin{figure}[!tb]
% 	\centering
% 	\includegraphics[width=0.8\columnwidth]{figs/decoupled_planning/framework_overview.pdf}
% 	\caption[An overview of the perception, planning and control framework]{
% 	Overview of our decoupled motion and foothold planning framework~\citep{Mastalli2015,Winkler2015}.
% 	The foothold planner first computes a sequence of body action and then selects the foothold locations $\mathbf{F}$.
% 	The foothold sequence is planned according the terrain information (i.e. terrain \revisedtext{cost-map} and heightmap).
% 	Then the motion planner uses the planned footholds to generate dynamic whole-body motions ($\mathbf{x}^d, \mathbf{\dot{x}}^d, \mathbf{\ddot{x}}^d$).
% 	Finally, the desired motion is compliantly executed by using a combination of feed-forward and feedback terms.
% 	(Figure modified from~\cite{Mastalli2017Thesis})}
% 	\label{fig:decoupled_framework}
% \end{figure}

\revisedtext{In our previous decoupled planning locomotion framework}~\citep{Winkler2015,Mastalli2015}, \revisedtext{the sequence of footholds was selected by computing an approximate body path}.
\revisedtext{It builds \textit{body-state graph} that quantifies the cost given a set of primitive actions towards a goal.}
Then, \revisedtext{it} chooses locally the locations of the footholds.
Finally, \revisedtext{it} generates a body trajectory that ensures dynamic stability and achieves the planned foothold sequence.
\revisedtext{For that, we used two fifth-order polynomials to describe the horizontal \gls{com} motion.
The stable horizontal motion is computed using a cart-table model.}
% \fref{fig:decoupled_framework} shows an overview of our decoupled motion and foothold planning framework.
For more details the reader can refer to~\citep{Winkler2015,Mastalli2015}.

%------------------------------ LOCOMOTION FRAMEWORK
% %%%%%%%%%%%%%%%%%%%%%%%%%%%%%%%%%%%%%%%%%%%%%%%%%%%%%%%%%%%%%%%%%%%%%%%%%%%%%%%
% 2345678901234567890123456789012345678901234567890123456789012345678901234567890
% 1         2         3         4         5         6         7         8
% %%%%%%%%%%%%%%%%%%%%%%%%%%%%%%%%%%%%%%%%%%%%%%%%%%%%%%%%%%%%%%%%%%%%%%%%%%%%%%%
% ******************************************************************************
% Locomotion framework
% ******************************************************************************
\section{Locomotion Framework}\label{sec:system}
\begin{figure*}[!tb]
	\centering
	\vspace{2mm}
	\includegraphics[width=1.0\textwidth]{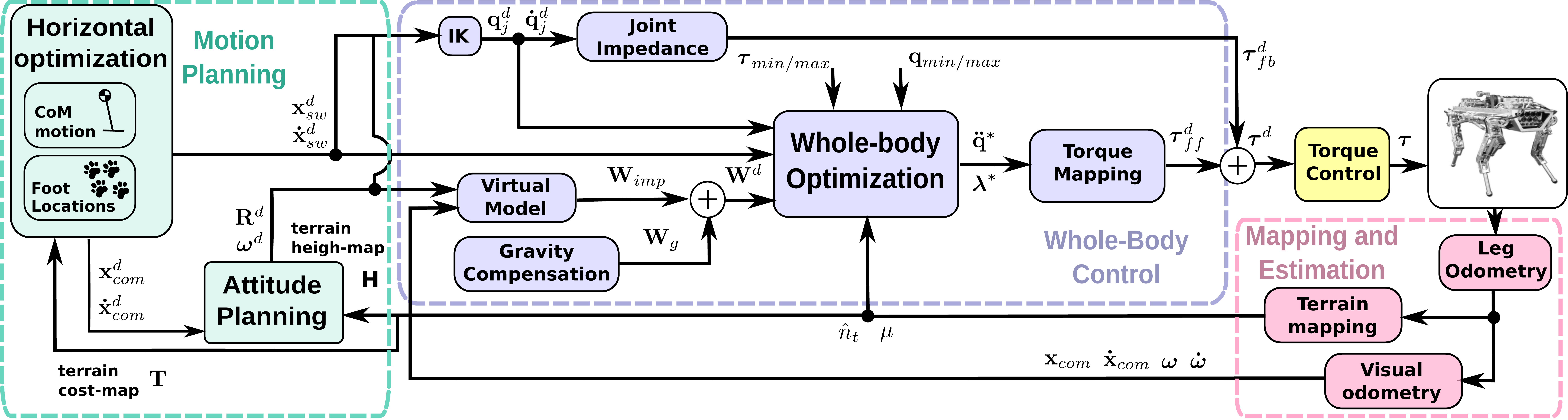}
    \caption{\revisedtext{Overview of the locomotion framework.
    Horizontal optimization computes simultaneously the \gls{com} and footholds $(\mathbf{x}_{com}^d,\mathbf{\dot{x}}_{com}^d,\mathbf{x}_{sw}^d,\mathbf{\dot{x}}_{sw}^d)$ given a terrain cost-map $\mathbf{T}$.
    The attitude planner adapts the trunk orientation $(\mathbf{R}^d,\boldsymbol{\omega}^d,\boldsymbol{\dot{\omega}}^d)$ given a terrain height-map $\mathbf{H}$.
    This results in a stable motion that is tracked by the whole-body controller.
	The virtual model allows us to compliantly track the desired \gls{com} motion, and the controller computes an instantaneous whole-body motion $(\vc{\ddot{q}}^*, \vc{\lambda}^*)$ that satisfies all robot constraints.
    The optimized motion is then mapped into desired feed-forward torques $\vc{\tau}^d$.
    To address unpredictable events, we use a joint impedance controller with low stiffness which tracks the desired joint commands $(\mathbf{q}^d_j,\mathbf{\dot{q}}_j^d)$.
    Finally, a state estimator provides an estimation of the trunk pose and twist $(\mathbf{x}_{com},\mathbf{\dot{x}}_{com},\boldsymbol{\omega},\boldsymbol{\dot{\omega}})$.
    It uses IMU and kinematics (leg odometry), and stereo vision (visual odometry).}}
	\label{fig:locomotionFramework}
\end{figure*}

\revisedtext{In this section, we give an overview of the main components of our locomotion framework (\sref{sec:framework}), after a quick description of the \gls{hyq} robot (\sref{sec:hyq}).}

\subsection{The HyQ robot}\label{sec:hyq}
\revisedtext{
\gls{hyq} is a \unit[85]{kg} hydraulically actuated quadruped robot~\citep{Semini2011}.
It is fully torque-controlled and equipped with precision joint encoders, a depth camera (Asus Xtion), a MultiSense SL sensor and an Inertial Measurement Unit (MicroStrain).
\gls{hyq} \revisedtext{measures approx.} \unit[1.0]{m}$\times$\unit[0.5]{m}$\times$\unit[0.98]{m} (length $\times$ width $\times$ height).
The leg \revisedtext{extension} length ranges from \unit[0.339-0.789]{m} and the hip-to-hip distance is \unit[0.75]{m} (in the sagittal plane).
It has two onboard computers: a Intel i5 processor with \gls{rt} Linux (Xenomai) patch, and a Intel i5 processor with Linux.
The Xenomai PC handles the low-level control (hydraulic-actuator control) at \unit[1]{kHz} and communicates with the proprioceptive sensors through EtherCAT boards.
Additionally, this PC runs the high-level (whole-body) controller at \unit[250]{Hz}.
Both \gls{rt} threads (i.e. low- and high- level controllers) communicate through shared memory.
On the other hand, the non-\gls{rt} PC processes the exteroceptive sensors to generate the terrain map and then compute the plans.
These motion plans are sent to the whole-body controller (i.e. the \gls{rt} PC) through a \gls{rt}-friendly communication.
}

\subsection{Framework components}\label{sec:framework}
\revisedtext{
Our locomotion framework is composed by three main modules: motion planning, whole-body control, and mapping and estimation (\fref{fig:locomotionFramework}).
The horizon optimization computes the \gls{com} motion and footholds to satisfy robot stability and to deal with terrain conditions (\sref{sec:preview_optimization}).
To handle terrain heights, our attitude planner adapts the trunk orientation as described in~\sref{subsec:trunk_modulation}.
We build onboard a terrain cost-map and height-map which are used by the horizontal and attitude planners, respectively (\sref{sec:terrain_costmap}).}

\revisedtext{
The whole-body controller has been designed to compliantly track motion plans (\sref{sec:execution}).
It consists of a virtual model, a joint impedance controller and a whole-body optimization.
The virtual model converts a desired motion into a desired wrench.
We additionally compensate for gravity to improve motion tracking.
Then, a whole-body optimization computes the joint accelerations and the contact forces that satisfy all the robot constraints (torque limits, kinematic limits and friction cone).
This output is mapped into desired feed-forward torque commands.
To address unpredictable events (e.g. slippage and contact instability), we include a joint impedance controller with low stiffness.
Finally, the torque commands are tracked by a torque controller~\citep{Boaventura2012b}.
}

\revisedtext{
The state estimation receives updates from proprioceptive sensors (IMU, torque sensors and encoders) as well as from exteroceptive sensors (stereo vision and LiDAR).
Faster updates (\unit[1]{kHz}) are obtained using leg odometry~\citep{Camurri2017}.
To correct drift, we fused at low frequencies visual odometry: optical flow and LiDAR registration~\citep{Nobili2017}.
The terrain mapping builds locally the cost-map and height-map using the depth camera (Asus Xtion).
With this, we obtain an accurate trunk position and terrain normals which are needed for the whole-body controller.
}

%------------------------------ COUPLED PLANNING
% %%%%%%%%%%%%%%%%%%%%%%%%%%%%%%%%%%%%%%%%%%%%%%%%%%%%%%%%%%%%%%%%%%%%%%%%%%%%%%%
% 2345678901234567890123456789012345678901234567890123456789012345678901234567890
% 1         2         3         4         5         6         7         8
% %%%%%%%%%%%%%%%%%%%%%%%%%%%%%%%%%%%%%%%%%%%%%%%%%%%%%%%%%%%%%%%%%%%%%%%%%%%%%%%
% ******************************************************************************
% Coupled motion and foothold planning
% ******************************************************************************
\section{\revisedtext{Motion planning}}\label{sec:coupled_planning}
\begin{figure}[!tb]
	\centering
	\vspace{2mm}
	\includegraphics[width=0.98\columnwidth]{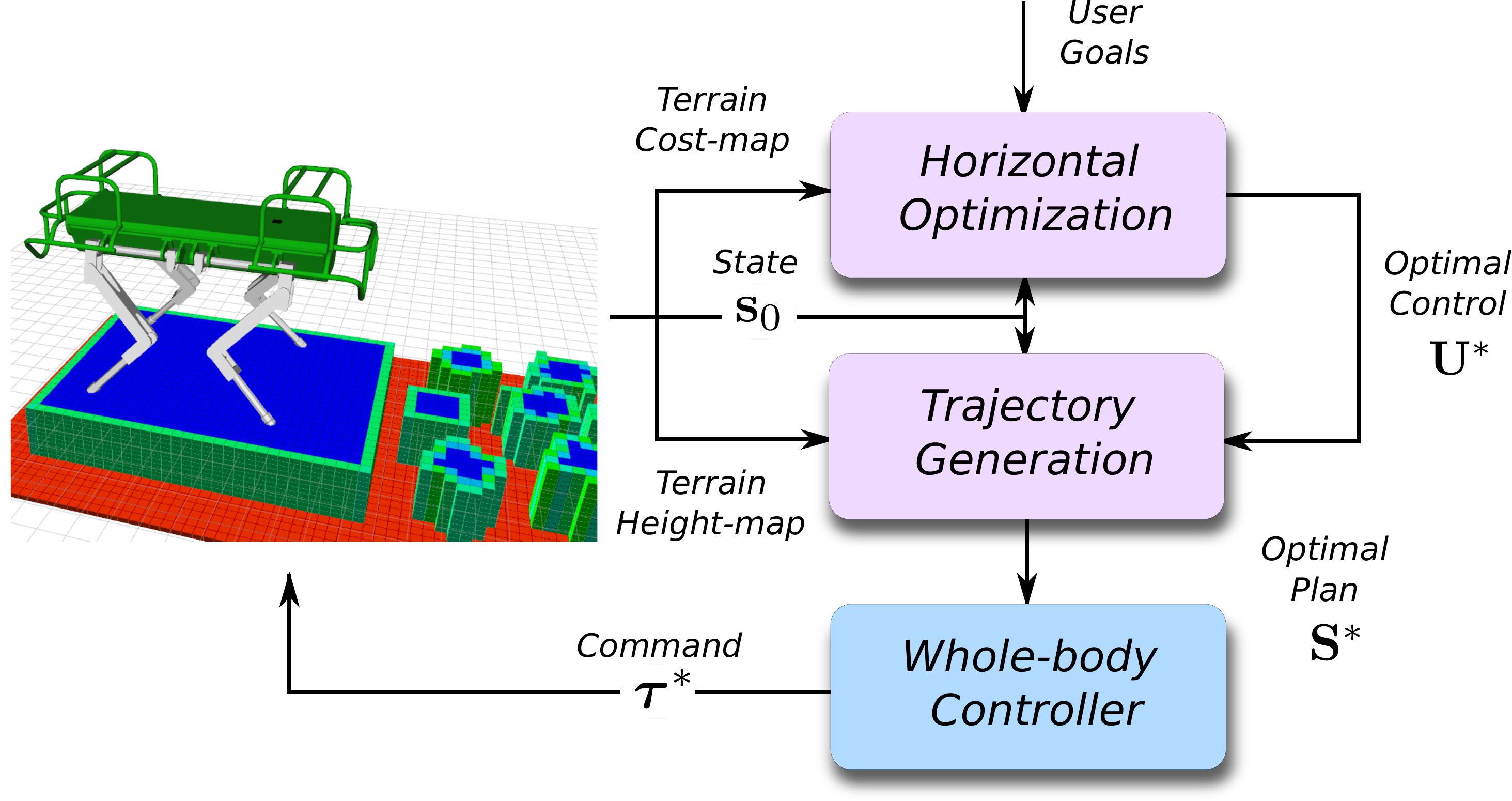}
	\caption[Overview of the trajectory optimization framework]{
	Overview of our coupled motion and foothold planning framework~\citep{Mastalli2017}.
	We compute offline an optimal sequence of control parameters $\vc{U}^*$ given the user's goals, the actual state $\vc{s}_0$ and the terrain cost-map.
	Given this optimal control sequence, we generate the optimal plan $\vc{S}^*$, that uses trunk attitude planning to adapt to the changes in the terrain elevation.
	Lastly, the whole-body controller calculates the joint torques \newtext{$\boldsymbol{\tau}$} that satisfy friction-cone constraints.
	(Figure from~\cite{Mastalli2017}.)}
	\label{fig:coupled_framework}
\end{figure}

We \revisedtext{consider} locomotion \revisedtext{to be} a coupled planning problem of \gls{com} motions and footholds (see \cref{fig:coupled_framework}).
First, we jointly generate the \gls{com} trajectory and the swing-leg trajectory using a sequence of parametric preview models and the terrain \revisedtext{height-map} (\Cref{sec:preview_generation}).
Then, \revisedtext{in~\Cref{sec:preview_optimization}, we optimize a sequence of control parameters, \revisedtext{given} the terrain cost-map, \revisedtext{which defines} a horizontal motion of the \gls{com}.}

\revisedtext{Key novelties}, with regards to previous work in~\citep{Mastalli2017}, \revisedtext{are the inclusion of} the terrain model in \revisedtext{the} 
optimization problem, \revisedtext{ as a} duality cost-constraint\revisedtext{, and the  development of  an attitude planning method.}
\revisedtext{The} terrain model allows us to navigate in various terrain conditions without the need \revisedtext{for} re-tuning.
\revisedtext{The attitude planner allows the robot to maintain stability in terrain with different elevations.}
\newtext{\revisedtext{Our} coupled planning approach allows \revisedtext{us} to optimize step timing and \revisedtext{to exploit the simplified} dynamics for foothold selection, an important improvement from our previous above-mentioned decoupled planning method.}

\subsection{Trajectory generation}\label{sec:preview_generation}
We generate the horizontal \gls{com} trajectory and the 2D foothold locations using a sequence of low-dimensional preview models (\sref{sec:horizontal_motion}).
\revisedtext{An optimization will provide a } sequence of control parameters \revisedtext{ for these models, that will form the horizontal \gls{com} trajectory. Not specifying the vertical motion for the \gls{com} allows us} to decouple the \gls{com} and trunk attitude planning.
\newtext{\revisedtext{In a second step, to achieve} dynamic adaptation to changes in the terrain elevation, we proposed a novel approach based on} the maximum allowed angular accelerations (\sref{subsec:trunk_modulation}).

\subsubsection{Preview model}
Preview models are low-dimensional (reduced) representations that are useful to describe and capture different locomotion behaviors, such as walking and trotting, and provide an overview of the motion~\citep{Mordatch2010, Kajita2003}.
With a reduced model we can still generate complex locomotion behaviors and their transitions; furthermore, we can integrate it with reactive control techniques.
In the literature, different models that capture the legged locomotion dynamics such as point-mass, inverted pendulum, cart-table, or contact wrench have been studied by~\cite{Full1999,Orin2013}.

\begin{figure}[tb]
	\centering
 	\includegraphics[width=0.9\columnwidth]{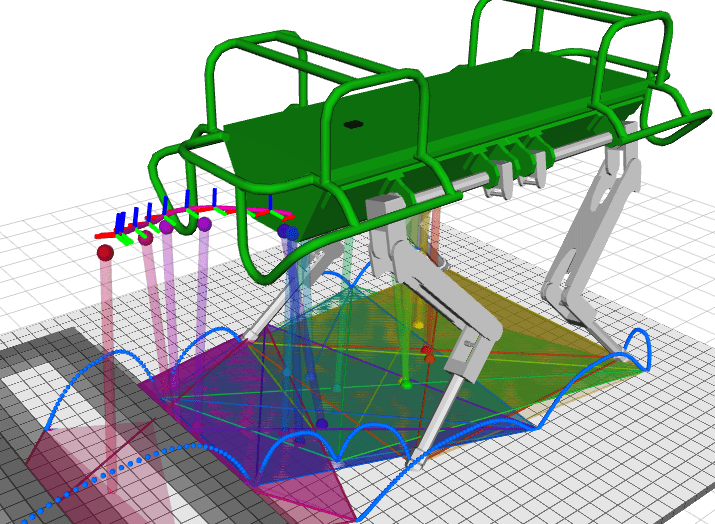}
 	\caption[A trajectory obtained from a low-dimensional model given a sequence of optimized control parameters.]{
	  A trajectory obtained from a low-dimensional model given a sequence of optimized control parameters.
	  The colored spheres represent the \gls{com} and \gls{cop} positions of the terminal states of each motion phase.
	  The \gls{cop} spheres lie inside the support polygon (same color is used).
	  Note that color indicates the phase (from yellow to red), \revisedtext{and the control parameters are computed from the terrain cost-map in grey-scale}.
	  The trunk adaptation is based on the estimated support planes in each phase.
	  Since the control parameters are expressed in the horizontal frame \revisedtext{(frame that coincides with the base frame but aligned with gravity)}, the horizontal \gls{com} trajectories and the trunk attitude are decoupled.
 	 (Figure from~\cite{Mastalli2017})}
	\label{fig:preview_generation}
\end{figure}

Our \revisedtext{cart-table with flywheel} model \revisedtext{(preview model)} \revisedtext{allows us to} decouple the \gls{com} motion from the trunk attitude\footnote{In this work, with \textit{trunk attitude} we refer to roll and pitch only.} (\cref{fig:preview_generation}).
\revisedtext{
To do not affect the \gls{cop} stability condition, we need to keep the \gls{cmp} inside the support region.
With the cart-table model, the horizontal optimization computes a sequence of control that keeps --within a safety margin-- the \gls{cop} inside the support polygon.
The attitude planner corrects the robot orientation in such a way that the \gls{cmp} position stays inside the support region.
This is possible due to the flywheel model allows us to predict the \gls{cmp} position given the trunk angular acceleration.}
\revisedtext{Note that} high centroidal moments (e.g. due to high trunk angular acceleration) can hamper the \gls{cop} stability condition e.g. causing that the \gls{cmp} moves out of the support polygon~\citep{Popovic2005} and making the robot losing its capability to balance.
% \newtext{Despite that, a safety margin in the \gls{cop} can be used to identify the maximum allowed moments applied to the \gls{com}, and by consequence to determining the maximum angular acceleration.}

\paragraph{\revisedtext{Horizontal} \gls{com} motion}\label{sec:horizontal_motion}
In our previous work~\citep{Winkler2015}, \newtext{we have observed that, in each locomotion phase, the \gls{cop} has an approximately linear displacement (see Fig. 8 from~\citep{Winkler2015})}\revisedtext{, i.e.:}
\begin{equation}
	\vc{p}^H(t) = \vc{p}^H_0 + \frac{\delta\vc{p}^H}{T}t,
	\label{eq:linear_cop}
\end{equation}
\revisedtext{where} $\vc{p}^H=(x^H,y^H)\in\mathbb{R}^2$ is the horizontal \gls{cop} position, $\delta\vc{p}^H\in\mathbb{R}^{2}$ the horizontal \gls{cop} displacement and $T$ is the phase duration.
The $(\cdot)^H$ apex means that the vectors are expressed in the horizontal frame.
\newtext{\revisedtext{As shown in~\citep{Mordatch2010}}, it possible\revisedtext{, using a cart-table model,} to derive an analytical expression for the horizontal \gls{com} trajectory\footnote{\revisedtext{The \gls{com} motion expressed in the horizontal frame. The horizontal frame coincides with base frame but aligned with gravity.}} whenever it is \revisedtext{assumed} a linear displacement of the \gls{cop}}:
\begin{eqnarray}
	\vc{x}^H(t) = \boldsymbol{\beta}_1 e^{\omega t} + \boldsymbol{\beta}_2
	e^{-\omega t} + \vc{p}^H_0 + \frac{\delta\vc{p}^H}{T}t,
	\label{eq:preview_model}
\end{eqnarray}
where the model coefficients $\boldsymbol{\beta}_{1,2}\in\mathbb{R}^{2}$ depend on the actual state $\vc{s}_0$ (horizontal \gls{com} position $\vc{x}^H_0\in\mathbb{R}^{2}$ and velocity $\dot{\vc{x}}^H_0\in\mathbb{R}^{2}$, and \gls{cop} position), the \gls{com} height $h$, the phase duration $T$, and the horizontal \gls{cop} displacement $\delta{\vc{p}}$:
\[\boldsymbol{\beta}_1 =
(\vc{x}^H_0-\vc{p}^H_0)/2 + (\dot{\vc{x}}^H_0 T -\delta{\vc{p}}^H)/(2\omega
T),\] \[\boldsymbol{\beta}_2 = (\vc{x}^H_0-\vc{p}^H_0)/2 - (\dot{\vc{x}}^H_0 T
-\delta{\vc{p}}^H)/(2\omega T),\] \revisedtext{with} $\omega = \sqrt{g/h}$ and $g$ is the gravity acceleration.

\paragraph{Trunk attitude}\label{subsec:trunk_modulation}
\revisedtext{The robot requires to adapt its trunk attitude when} the terrain elevation varies.
As \newtext{naive} \revisedtext{heuristic}, \revisedtext{we could always try to align} the trunk with respect to the estimated support plane\footnote{\revisedtext{A course estimation of the supporting plane can be obtained fitting an averaging plane across the feet in stance, with an update at each touch-down.}}.
\newtext{\revisedtext{However, blindly following the heuristics can lead to big changes in the orientation.
In this case big moments} might move the \gls{cmp} out of the support polygon, \revisedtext{thus} invalidating the \gls{cop} condition for stability \revisedtext{and making the robot lose control authority}.
\revisedtext{To address this issue, we propose to limit the attitude adaptation by introducing bounds that can guarantee the motion stability.}
\revisedtext{For that, we observe that,} for a cart-table with flywheel model, \revisedtext{the \gls{cmp} $\vc{m}\in\mathbb{R}^{3}$ is linked to the \gls{cop} $\vc{p}\in\mathbb{R}^{3}$} ~\citep{Koolen2012,Popovic2005} as}:
\begin{equation}
	\newtext{\vc{m} = \vc{p} + \Delta}
\end{equation}
\newtext{where  $\Delta$ is the shift resulting from applying moment to the \gls{com}}:
\begin{align}
	\label{eq:com_torque}
	\Delta_x& =  \tau_{{com}_y}/mg, \\ \nonumber
	\Delta_y& = \newtext{-} \tau_{{com}_x}/mg,
\end{align}
\revisedtext{and} $\tau_{com_y}$, $\tau_{com_x}$ \revisedtext{are} the horizontal components of the moment about the \gls{com}.
\newtext{\revisedtext{Then,} thanks to the} simplified flywheel model we can also link these moments to \newtext{angular accelerations of} the \newtext{\gls{com}, i.e. $\boldsymbol{\tau}_{com} = \mathcal{I}\dot{\boldsymbol{\omega}}$.
Re-writing \eref{eq:com_torque} in vectorial form we have}:
\begin{equation}
	\newtext{\Delta = \frac{\mathcal{I}\dot{\boldsymbol{\omega}}\times{\vc{\hat{e}}_z}}{mg}.}
\label{eq:margin}
\end{equation}
where $\mathcal{I}\in\mathbb{R}^{3\times 3}$ is the time-invariant inertial tensor approximation \revisedtext{(e.g. for a default joint configuration)} of the centroidal inertia matrix of the robot, and \newtext{$\vc{\hat{e}}_z$ is z basis vector of the inertial frame.}

\revisedtext{Under the flywheel assumption, we can keep} \newtext{the \gls{cmp} inside the support polygon} by limiting the angular acceleration \newtext{of the trunk \revisedtext{to} $\dot{\boldsymbol{\omega}}_{max}$, \revisedtext{where this value is computed from} a safety margin $r$ defined} \newtext{in} the \newtext{trajectory} optimization (\newtext{see} \sref{subsec:constraints}) as:
\begin{equation}\label{eq:max_ang_acc}
	r = \newtext{\Big\| \frac{(\mathcal{I}\dot{\boldsymbol{\omega}}_{max})\times{\vc{\hat{e}}_z}}{mg}\Big\|.}
\end{equation}

\newtext{With this}, we \newtext{can} adapt the trunk attitude \revisedtext{without affecting} the \newtext{stability of the robot.
\revisedtext{To summarize} we want to align the trunk \revisedtext{to} the estimated support plane.
However, \revisedtext{the level of alignment is limited by the maximum} allowed angular accelerations, in the frontal and the transverse plane (i.e. $\dot{\omega}_x$ and $\dot{\omega}_y$), computed from \eref{eq:max_ang_acc} given a \revisedtext{user-}defined $r$ margin}.

We employ cubic polynomial splines to describe the trunk attitude motion (frontal and transverse).
The attitude adaptation can \revisedtext{be achieved in more than one} phase.
\newtext{Indeed, the} required angular displacement \newtext{\revisedtext{may not be} possible without exceeding the allowed angular accelerations}.

\paragraph{\newtext{Trunk height}}
\newtext{We do not consider vertical dynamics during the \gls{com} generation, instead we assume that the trunk height is constant \revisedtext{throughout} the motion.
\revisedtext{For} non-coplanar footholds, we use an estimate of the support area \revisedtext{which provides the height of the \gls{cop} point}.
With this, the sequence of parameters expressed in the horizontal plane are still valid, i.e. they do not affect the robot stability.}

\subsubsection{Preview schedule}\label{subsec:preview}
\begin{figure}
	\centering
 	\includegraphics[width=0.9\columnwidth]{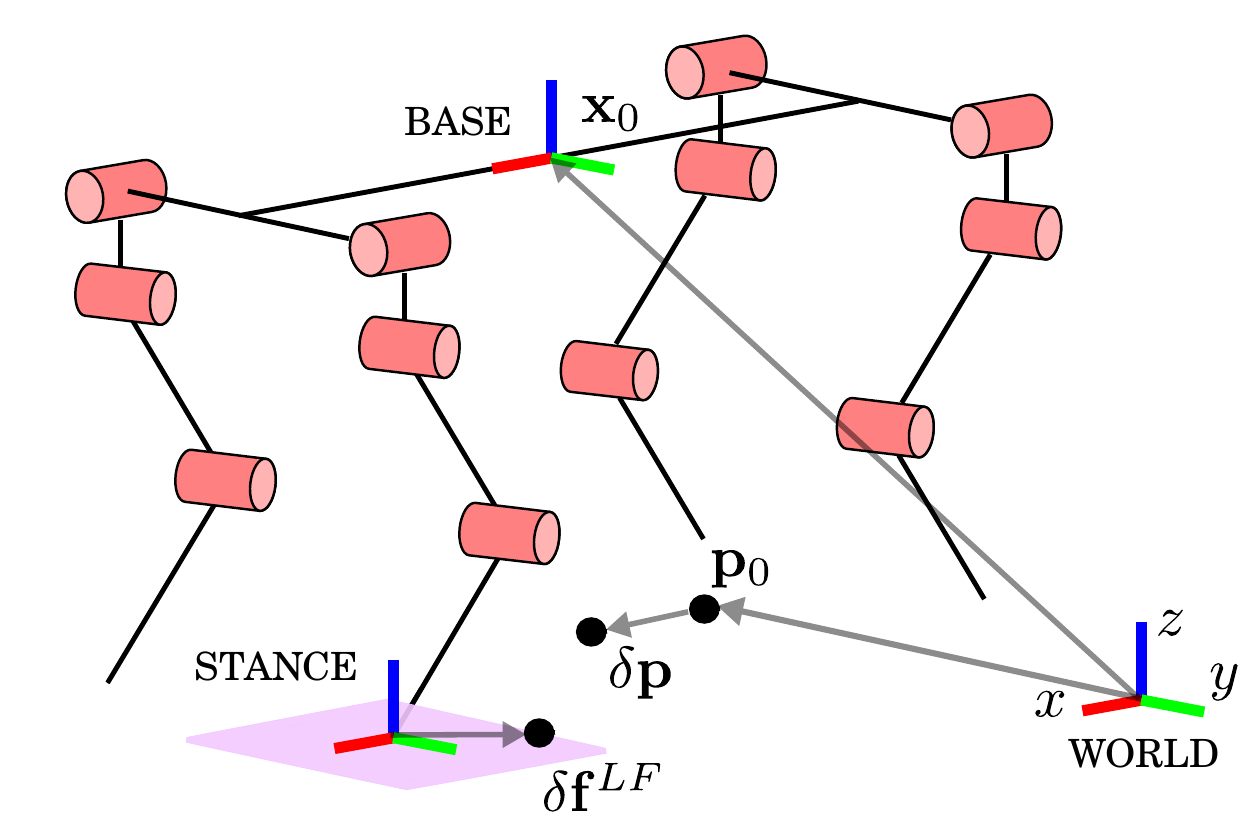}
 	\caption[Sketch of different variables and frames used in our optimization]{
 	Sketch of different variables and frames used in our optimization.
	 The foot-shift $\delta\vc{f}^{LF}$ is described w.r.t. the stance frame, its bounds are defined by the foothold	region (the pink rectangle).
	 The stance frame is calculated from the default posture and expressed w.r.t. the base frame.
	 (Figure modified from~\cite{Mastalli2017}.)}
	\label{fig:optimization_variable_sketch}
\end{figure}

Describing legged locomotion \revisedtext{is possible using} a sequence of different preview models --- a preview schedule.
Using this, \revisedtext{we} can \revisedtext{define} different foothold sequences by enabling or disabling different phases in our optimization process, \revisedtext{i.e. with phase duration equals zero $(T_i = 0)$}.
In the preview schedule, we build up a sequence of control parameters $\mathbf{U}$ \revisedtext{from stance $\vc{u}^{st}$ and foot-swing $\vc{u}^{sw}_i$ phases as follows:}
\begin{equation}
	\vc{U} = \mat{\vc{u}^{st/sw}_1 & \cdots & \vc{u}^{st/sw}_n},
\end{equation}
\revisedtext{in which the phases are defined as $\vc{u}^{st}_i = \mat{T_i & \delta\vc{p}^H_i}$ \textit{stance phase} where all legs are on the ground and $\vc{u}^{sw}_i = \mat{T_i &\delta\vc{p}^H_i & \delta\vc{f}^l_i}$, \textit{swing phase}, where at least one leg is in swing. $n$ is the number of phases, $l$ is the swing foot, $T_i$ is the phase duration and $\delta\vc{f}_i^l$ is the relative foothold location (i.e. \textit{foot-shift}) described w.r.t. the stance frame (\cref{fig:optimization_variable_sketch})}.
The stance frame is computed from a default posture of the robot.
\revisedtext{Note that the foothold locations do not affect the \gls{com} dynamics since our model neglects the leg masses and the angular dynamics (cart-table model). In addition, due to the fact that the foot location is an optimization variable that affects the shape of the polygon used as stability constraint for the \gls{zmp}, we have the product of  two optimization variables, hence the problem becomes non-linear.} 

\newtext{The horizontal \gls{com} trajectory is computed from a sequence of phase duration and \gls{cop} displacements $\{T_i,\delta\vc{p}_i^H\}$, and its height is kept constant according to the estimated support polygon.
\revisedtext{Note that the} trunk attitude adaption depends on the safety margin and the footstep heights (computed from the terrain height-map).
In short, the \gls{com} trajectory can be represented as follows:}
\begin{equation}\label{eq:preview_schedule}
	\newtext{\vc{S} = \{\vc{s}_1,\cdots,\vc{s}_N\} = \vc{f}(\vc{s}_0, \vc{U}, r)}
\end{equation}
\newtext{where the preview state \revisedtext{$\vc{s} = \mat{\vc{x} & \dot{\vc{x}} & \vc{p} & \vc{\sigma}}$} is defined by the \gls{com} position and velocity $(\vc{x},\dot{\vc{x}})$, \gls{cop} position $\vc{p}$ and the stance support region $\vc{\sigma}$, \revisedtext{which is defined by the active feet described by $\mathbf{U}$}.
For simplicity, we have described only the initial preview states of each phase in~\eref{eq:preview_schedule}.
\revisedtext{It} is possible to recover any state because $\vc{f}(\cdot)$ describes the time-continuous \gls{com} dynamics.}
Using preview models it is important to reduce the number of decision variables (through control parameters). 

We are focused on finding a minimum and safer sequence of footsteps given a certain terrain condition. Therefore we need to include the terrain model.
A terrain model often is non-convex \revisedtext{and not necessary differentiable}.
\revisedtext{If incorporated in an optimization problem this can create plenty of local minima, then stochastic optimization is a promising choice as a solver (more details in~\sref{sec:preview_optimization})}.
\revisedtext{In the following we will provide the description of the terrain model.}

\subsection{\newtext{Terrain cost-map}}\label{sec:terrain_costmap}
\newtext{\revisedtext{Our terrain model is represented by the \textit{terrain cost-map}.
This quantifies how desirable it is to place a foot at a specific location.}
The cost value for each pixel in the map is computed using geometric terrain features such as \textit{height deviation}, \textit{slope} and \textit{curvature}~\citep{Kolter2009}.
These values are computed as a weighted linear combination of the individual features  $T(x,y) = \vc{w}^T \vc{T}(x,y)$, where $\vc{w}$ and $\vc{T}(x,y)$ are the weights and feature cost values, respectively.
The total cost value is normalized, where 0 and 1 \revisedtext{represent the minimum and maximum risk to step in}, respectively.
The weight vector describes the importance of the different features.
Each feature is computed through piece-wise functions that resemble the log-barrier constraint \revisedtext{as described in the equations of~\sref{sec:slope},~\ref{sec:curvature}}.
With this, we have \revisedtext{extended} the off-the-shelf solver to \revisedtext{address} constrained problems (see~\sref{sec:preview_optimization} for more details).
Below the log-barrier function for each feature \revisedtext{is described}.}

\subsubsection{\newtext{Height deviation}}
\newtext{The height deviation cost penalizes footholds close to large drop-offs; for instance, this cost is \revisedtext{important} for crossing gaps or stepping stones.
In fact, staying far away from large drop-offs is beneficial because inaccuracies in the execution of footsteps can cause the robot to step into gaps or banned areas.
The height deviation feature $f_h$ is computed using the standard deviation around a defined neighborhood.}

\subsubsection{\newtext{Slope}}\label{sec:slope}
\newtext{The slope reflects the local surface normal in a neighborhood around the cell.
\revisedtext{The} normals are computed using \gls{pca} on the set of nearest neighbors.
A high slope value will increase the chance of slipping even in cases where the friction cone is considered, e.g. due to inaccuracies in the friction coefficient or estimated surface normal.
Slope cost increases for larger slope values, while small slopes have zero cost as they are approximately flat.
We consider the worst possible slope $s_{max}$ occurs when the terrain is very steep (approximately 70$^{\circ}$)\footnote{\newtext{We heuristically defined this value based on our experience with the \gls{hyq} robot and its geometry.}}.}

\newtext{We map the height deviation and slope features \revisedtext{$f$} into cost values through the following piece-wise function:
\[
T_f(x,y) =
\begin{cases}
  0  											   &	f\leq f_{flat} \\
 -\ln\Big(1-\frac{f(x,y)}{f_{max}-f_{flat}}\Big) &	f_{flat} < f < f_{max} \\
 T_{max}										   &	f\geq f_{max}
\end{cases}
\]
where $f_{flat}$ is a threshold that defines the flat conditions, $f_{max}$ the maximum allowed feature value, and $T_{max}$ is the maximum cost value.}
\revisedtext{Note that $f_{max}$ defines the barrier of the log function.}

\subsubsection{\newtext{Curvature}}\label{sec:curvature}
\newtext{The curvature describes the \revisedtext{contact} stability of a given foothold location.
For instance, terrain with mild curvature (curvature between $c = 6$ to $c = 9$) is preferable to flat terrain since it reduces the possibility of slipping, as it has a bowl-like structure.
Thus, the cost is equal to zero in those conditions.
On the other hand, high and low curvature values represent a narrow crack structure ($c > 9 = c_{\max}$) or edge structure
($c < -6 = c_{\min}$) in which the foot can get stuck in or can slip, respectively.
We use the following piece-wise function to compute the cost value from a curvature value $c$:
\[
T_c(x,y) =
\begin{cases}
  T_{max} - \ln\Big(\frac{c(x,y)-c_{\min}}{c_{\max}-c_{\min}}\Big)   &	c_{crack} < c < c_{mild}^- \\
  0 															   &    c_{mild}^- < c < c_{mild}^+ \\
  T_{max}  										   			 	   &	c\leq c_{crack}, \\
\end{cases}
\]
where $c_{crack} = -6$, $c_{mild}^- = 6$, $c_{mild}^+ = 9$ and $c_{\max} = 9$.
A description of different curvature values can be found in \cite{Kalakrishnan2010b}.}
\revisedtext{The barriers are defined by $c_{crack}$ and $c_{mild}$.}

\subsection{\revisedtext{Horizontal trajectory} optimization}\label{sec:preview_optimization}
The trajectory optimization computes an optimal sequence of control parameters $\vc{U}^*$ used for the generation of the \revisedtext{horizontal trajectories for the \gls{com}} (\Cref{sec:preview_generation}).
\revisedtext{We compute the entire plan by solving a finite-horizon trajectory optimization problem for each phase (similarly to a receding horizon strategy).}
The horizon is described by a predefined number of preview schedules $N$ with $n$ \revisedtext{phases} (e.g. our locomotion cycle (schedule) has 6 \revisedtext{phases}).
\revisedtext{Our method presents several advantages to address} challenging terrain locomotion.
It enables \revisedtext{the robot} to generate desired behaviors that anticipate future terrain conditions, \revisedtext{which} results in smoother transitions between phases.
\revisedtext{Note that the optimal solution $\vc{U}^*$ is defined as explained in~\sref{subsec:preview}.}

\newtext{Compared \revisedtext{with}~\cite{Mordatch2010}, our trajectory optimization method 1) uses a terrain cost-map model for foothold selection, 2) defines non-linear inequalities constraints for the \gls{cop} position, and 3) guarantees the robot stability against changes in the terrain elevation.
Additionally\revisedtext{, in contrast to ~\cite{Mordatch2010},} we have defined a single cost function that tracks desired walking velocities\revisedtext{, without enforcing the tracking of a specific step time and distance}.}

\subsubsection{\revisedtext{Problem formulation}}
Given an initial state $\vc{s}_0$, we optimize a sequence of control parameters inside a predefined horizon, and apply only the optimal control of the current phase.
\revisedtext{Given the desired user commands (trunk velocities), a} sequence of control parameters $\vc{U}^*$ \revisedtext{are computed solving} an unconstrained optimization problem:
\begin{equation}
	\vc{U}^* = \argmin_{\vc{U}} \sum_j \omega_j g_j(\newtext{\vc{s}_0,\vc{U},r}).
	\label{eq:preview_optimization}
\end{equation}
We solve this trajectory optimization problem using the \gls{cmaes}~\citep{Hansen2014}.
\gls{cmaes} is capable of handling optimization problems that have multiple local minima \newtext{and discontinuous gradients}.
\revisedtext{An important feature since the terrain \revisedtext{cost-map} introduces multiple local minima and gradient discontinuity}.
We use soft-constraints as these provide the required freedom to search in the landscape of our optimization problem.
The cost functions or soft-constraints $\newtext{g_j(\vc{s}_0,\vc{U},r)}$ describe:
\textit{1}) the user command as desired walking velocity and travel direction, \textit{2}) the energy, \textit{3}) the terrain cost, \textit{4}) a soft-constraint to ensure stability (i.e. the \gls{cop} condition), and \textit{5}) \revisedtext{a soft-constraint that ensures the coupling between} the horizontal and vertical dynamics, \revisedtext{where the horizontal dynamics are described by~\eref{eq:preview_model}}.
\subsubsection{Cost functions}\label{subsec:cost}
We \revisedtext{use the average walking velocity to track the user velocity command.}
We evaluate the desired velocity command for the entire planning horizon $Nn$ as follows:
\begin{eqnarray}
	g_{velocity} = \bigg(\dot{\vc{x}}^H_{desired} - \frac{\vc{x}^H_{Nn} -
	\vc{x}^H_0}{\sum_{i=1}^{Nn} \newtext{T_i}}\bigg)^2,
\end{eqnarray}
where $\dot{\vc{x}}^H_{desired}\in\Rnum^2$ is the desired horizontal velocity, $\vc{x}^H_{Nn}$ is the terminal \gls{com} position, $\vc{x}^H_{0}$ is the actual \gls{com} position, and $T_i$ is the duration of $i^{th}$ phase.
Note that \revisedtext{$\vc{x}^H_{Nn}$ is} the latest state that we consider in the planning horizon.

\revisedtext{We use an estimated measure of} the energy \revisedtext{needed to move} the robot from one place to another \revisedtext{which we call the \gls{mcot}}.
Minimizing the \gls{mcot} reduces the energy consumption for \newtext{traversing} a \newtext{given} terrain.
Since joint torques and velocities \revisedtext{are not available} in our optimization, we \newtext{approximate} the \revisedtext{robot} kinetic energy \revisedtext{with a single point-mass system} (i.e. $\mathcal{K} = \frac{1}{2}m\dot{\vc{x}}^2$).
Thus, we compute the total cost along the phases by:
\begin{equation}
	g_{elc} = \sum_{i=1}^{Nn} ELC(\dot{\vc{x}}),
\end{equation}
where $ELC(\dot{\vc{x}}) \triangleq \frac{\mathcal{K}}{mgd}$ with $d$ \revisedtext{equal} to the travel distance in the $xy$ plane.

\subsubsection{\newtext{Soft-constraints}}
\label{subsec:constraints}
\begin{figure}[!tb]
	\centering
	\vspace{2mm}
 	\includegraphics[width=0.9\columnwidth]{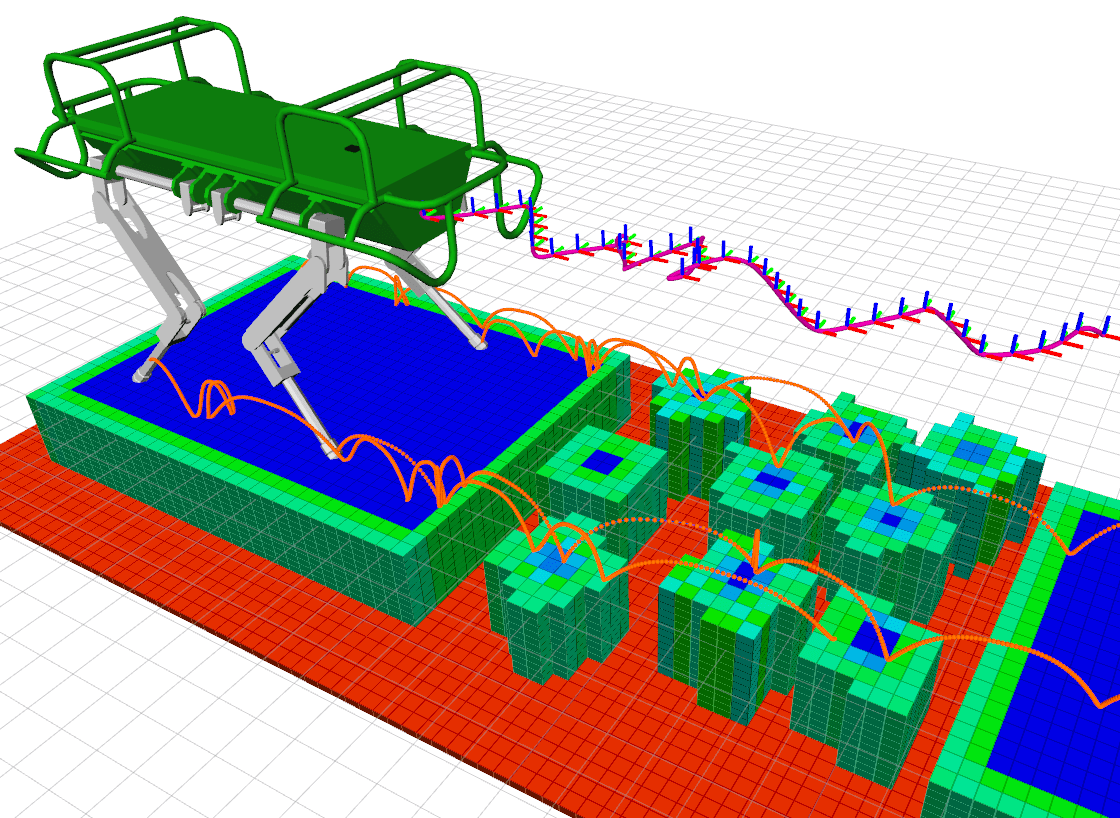}
 	\caption[A \revisedtext{cost-map} allows the robot to negotiate different terrain conditions
 	while following the desired user commands]{
	 A \revisedtext{cost-map} allows the robot to negotiate different terrain conditions while following the desired user commands.
	 The \revisedtext{cost-map} is computed from onboard sensors as described in~\Cref{sec:terrain_costmap}.
	 The cost values are continuous and represented in color scale, where blue is the minimum and red is the maximum cost. (Figure from~\cite{Mastalli2017}.)}
	\label{fig:preview_optimization}
\end{figure}

To \revisedtext{negotiate} different terrains (\cref{fig:preview_optimization}), we compute \revisedtext{onboard} the \revisedtext{cost-map} as described in~\Cref{sec:terrain_costmap}.
Thus, given a foot-shift and \gls{com} position, we \revisedtext{can obtain the cost at the correspondent foothold location $(x,y)$} as:
\begin{equation}\label{eq:terrain_weight}
	g_{terrain} = \vc{w}^T \vc{T}(x,y),
\end{equation}
\revisedtext{where $\vc{w}$ and $\vc{T}(x,y)$ are the weights and the vector of cost values of every feature at the location $(x,y)$, respectively.}
\newtext{Note that each feature is computed as explained in~\sref{sec:terrain_costmap}, and \revisedtext{they define log-barrier constraints on the terrain.}}
We use a cell grid resolution of \unit[2]{cm}, a half of the robot's foot size.
As in~\citep{Mastalli2015}, we demonstrated that this coarse map is a good trade-off in terms of computation time and information resolution for foothold selection.
We cannot guarantee convexity in the terrain costmap, which has to be considered in our optimization process.

As mentioned in~\sref{subsec:trunk_modulation}\revisedtext{, in order to ensure a certain motion freedom for the control of the attitude, we keep the \gls{cop} trajectory  inside a  polygon that it is shrunk by a margin $r$ with respect to the support polygon.}
We use a set of non-linear inequality constraints to describe the \newtext{shrunk} support region:
\begin{equation}\label{eq:support}
	\vc{L}(\vc{\sigma}\newtext{,r})^T \mat{\vc{p} \\ 1} > \vc{0},
\end{equation}
where $\vc{L}(\cdot)\in\mathbb{R}^{l\times3}$ are the coefficients of the $l$ lines, $\vc{\sigma}$ the support region defined from the  foothold locations, and $\vc{p}$ the \gls{cop} position.
\revisedtext{Note that it is a nonlinear constraint as we include} the foothold positions as decision variables.

\revisedtext{Due to the cart-table model assumes a constant height, the consistency between the \gls{com} and \gls{cop} motion is ensured by imposing the following soft-constraint:}
\begin{equation}\label{eq:coupling}
	h = \|\vc{x} - \vc{p}\|
\end{equation}
where \newtext{$h$ is the \revisedtext{cart-table height that describes the default height of the robot},} \revisedtext{and} $\vc{x}$ and $\vc{p}$ are the \gls{com} and \gls{cop} positions, respectively.
\revisedtext{This soft-constraint penalizes the artificial increment of the \gls{com} horizontal position that appears when the decoupling with the vertical motion becomes inaccurate (see~\eref{eq:preview_model}).}

We impose both soft-constraints \revisedtext{(i.e.~\eref{eq:support},~(\ref{eq:coupling}))} only in the initial and terminal state of each phase.
\revisedtext{This is sufficient because the stability and the coupling} will be guaranteed in the entire phase \revisedtext{too}.
\revisedtext{Note that, for the stability constraint, the support polygon remains a convex hull as the possible foothold locations cannot cross its geometric center.}
We ensure this by limiting the foothold search region, i.e. by bounding the foot-shift (see~\cref{fig:optimization_variable_sketch}).
These soft-constraints are described \revisedtext{using quadratic penalization}.

%------------------------------ EXECUTION
% %%%%%%%%%%%%%%%%%%%%%%%%%%%%%%%%%%%%%%%%%%%%%%%%%%%%%%%%%%%%%%%%%%%%%%%%%%%%%%%
% 2345678901234567890123456789012345678901234567890123456789012345678901234567890
% 1         2         3         4         5         6         7         8
% %%%%%%%%%%%%%%%%%%%%%%%%%%%%%%%%%%%%%%%%%%%%%%%%%%%%%%%%%%%%%%%%%%%%%%%%%%%%%%%
% ******************************************************************************
% Whole-body controller
% ******************************************************************************
\section{Whole-body controller}\label{sec:execution}
The \newtext{tracking of the reference trajectories for the \gls{com} ($\vc{x}_{com}^d, \vc{\dot{x}}_{com}^d, \vc{\ddot{x}}_{com}^d$), the trunk orientation ($\pmb{R}^d, \pmb{\omega}^d, \pmb{\dot{\omega}}^d$) and the swing motions ($\vc{x}_{sw}^d, \vc{\dot{x}}_{sw}^d$) is ensured by a whole-body controller (i.e. \textit{trunk controller}).}
This computes the feed-forward joint torques \newtext{$\vc{\tau}_{ff}^d$} necessary to achieve a desired motion without violating friction, torques \newtext{($\vc{\tau}_{max,min}$)} or kinematic limits \newtext{($\vc{q}_{max,min}$).} 
To fulfill these additional constraints we exploit the redundancy in the mapping between the joint space ($\in\Rnum^n$) and the body task ($\in\Rnum^6$).
To address unpredictable events (e.g. limit foot divergence in \revisedtext{the} case of slippage on an unknown surface), an impedance controller computes in parallel the feedback joint torques $\vc{\tau}_{fb}$ from the desired joint motion $(\mathbf{q}^d_j, \dot{\mathbf{q}}^d_j)$.
This controller receives position/velocity set-points that are consistent with the body motion to prevent conflicts with the trunk controller.
In nominal operations the biggest contribution is generated by the feed-forward torques, i.e. by the trunk controller.

\newtext{This controller has been previously drafted in~\citep{Aceituno2017b} and subsequently presented in detail in~\citep{FahmiMastalli2019}.
\revisedtext{Our controller extends previous work on whole-body control, in particular~\citep{Ott2011,Henze2017}.}
In this section we briefly summarize its main characteristics.
We cast the controller as an optimization problem, in which, by incorporating the \textit{full} dynamics of the legged robot, all of its \gls{dofs} are exploited to spread the desired motion tasks globally to all the joints.}

\newtext{Although the usage of a reduced model (e.g. a centroidal model) can be convenient for planning purposes, in control, it is important to consider the dynamics of all the joints when dealing with dynamic motions (as \revisedtext{shown} in~\sref{sec:results}).}
\newtext{In these cases, the effect of the leg dynamics is no longer negligible and must be considered to achieve good tracking.} 

\newtext{With this whole-body controller, the robot achieves faster dynamic motions in real-time, \revisedtext{see~\citep{FahmiMastalli2019}}, \revisedtext{when} compared with our previous quasi-static controller~\citep{Focchi2017a}.
The block diagram of the trunk controller is \revisedtext{shown} in~\fref{fig:locomotionFramework}.
A virtual model generates the reference centroidal wrench $\vc{W}_{imp}$ necessary to track the reference trajectories. 
\revisedtext{The} problem \revisedtext{is formulated} as a \gls{qp} with the generalized accelerations and contact forces as decision variables, i.e.
$\mathbf{x}=[\ddot{\mathbf{q}}^T,\boldsymbol{\lambda}^T]^T\in\mathbb{R}^{6+n+3n_l}$ where $n_l$ is the number of end-effectors in contact:}
\revisedtext{
\begin{equation}
    \label{eq:qp}
    \begin{aligned}
        \mathbf{x}^* = \argmin_{\mathbf{x}=(\ddot{\mathbf{q}},\boldsymbol{\lambda})}
        &\hspace{-2.em}
        & & \hspace{-0.75em}\Vert \vc{W}_{com} - \vc{W}_{com}^d \Vert^2_{\vc{Q}}+\|\mathbf{x}\|_{\vc{R}}^2\hspace{-8.em}&\\
        & \hspace{-3em}\textrm{s.t.}
        & & \hspace{-2em}\mathbf{M}\ddot{\mathbf{q}} + \mathbf{h} = \mathbf{S}\boldsymbol{\tau} + \mathbf{J}_c^T\boldsymbol{\lambda}, &\textrm{(dynamics)}\\
        & & & \hspace{-2em}\mathbf{J}_c\ddot{\mathbf{q}} + \dot{\mathbf{J}}_c\dot{\mathbf{q}} = \mathbf{0}, &\textrm{(stance)}\\
        & & & \hspace{-2em}\mathbf{R}\boldsymbol{\lambda} \leq \mathbf{\mathbf{r}}, &\textrm{(friction)}\\
        & & & \hspace{-2em}\mathbf{\bar{\ddot{q}}} \leq \mathbf{\ddot{q}} \leq \mathbf{\underline{\ddot{q}}}, &\textrm{(kinematics)}\\
        & & & \hspace{-2em}\mathbf{\bar{\boldsymbol{\tau}}} \leq \mathbf{M}\ddot{\mathbf{q}}+\mathbf{h}-\mathbf{J}_c^T\boldsymbol{\lambda} \leq \mathbf{\underline{\boldsymbol{\tau}}}, &\textrm{(torque)}
    \end{aligned}
\end{equation}
where $\mathbf{M}$ is the joint-space inertial matrix, $\mathbf{J}_c$ is the contact Jacobian, $\mathbf{h}$ is the force vector that accounts Coriolis, centrifugal, and gravitational forces, $(\mathbf{R},\mathbf{r})$ describe the linearized friction cone, $(\mathbf{\bar{\ddot{q}}},\mathbf{\underline{\ddot{q}}})$ are acceleration bounds defined given the current robot configuration~\cite{FahmiMastalli2019}, and $(\mathbf{\bar{\boldsymbol{\tau}}}, \mathbf{\underline{\boldsymbol{\tau}}})$ describe the torque limits.}

\newtext{The first term of the cost function~\eqref{eq:qp} penalizes the \textit{tracking} error at the wrench level, while the second one is a regularization factor to keep the solution bounded or to pursue additional criteria.
Both costs are quadratic-weighted terms.
All the constraints are linear: the equality constraints encode dynamic consistency, the stance condition and the swing task.
While the inequality constraints encode friction, torque, and kinematic limits.
Then the optimal acceleration and forces $\mathbf{x}^*$ are mapped into desired feed-forward joint torques
$\boldsymbol{\tau}_{ff}^d\in\mathbb{R}^n$ using the actuated part of the full dynamics.}
\newtext{Finally, the feed-forward torques $\boldsymbol{\tau}_{ff}^d$ are summed with the joint PD torques (i.e. feedback torques $\boldsymbol{\tau}_{fb}$) to form the desired torque command $\boldsymbol{\tau}^d$, which is sent to a low-level joint-torque controller.}

\newtext{A \textit{terrain mapping} module provides, as inputs to the whole-body controller, an estimate of the friction coefficient $\mu$ and of the normal to the terrain $\hat{n}_t$ at each contact location~\cite{Mastalli2019report}.
Finally a \textit{state estimation} module fuses inertial, visual and odometry information to get the current floating-base position and velocity w.r.t. the inertial frame \cite{Nobili2017}.}

%------------------------------ RESULTS
% %%%%%%%%%%%%%%%%%%%%%%%%%%%%%%%%%%%%%%%%%%%%%%%%%%%%%%%%%%%%%%%%%%%%%%%%%%%%%%%
% 2345678901234567890123456789012345678901234567890123456789012345678901234567890
% 1         2         3         4         5         6         7         8
% %%%%%%%%%%%%%%%%%%%%%%%%%%%%%%%%%%%%%%%%%%%%%%%%%%%%%%%%%%%%%%%%%%%%%%%%%%%%%%%
% ******************************************************************************
% Experimental Results
% ******************************************************************************
\section{Experimental Results}\label{sec:results}
To \revisedtext{understand the advantages of our locomotion framework}, we first compare the decoupled and coupled approaches in \revisedtext{different} challenging terrains
(e.g. stepping stones, pallet, stairs and gap).
We use as test-cases \revisedtext{for the comparison} the decoupled planner presented in~\citep{Winkler2015}.
After that, we \revisedtext{show} how the \revisedtext{modulation} of the trunk attitude \revisedtext{handles heights variations in the terrain}.
Subsequently, we analyze the effect of the terrain \revisedtext{cost-map} in our coupled planner.
We \revisedtext{study} how different weighting choices result in \revisedtext{different} behaviors without affecting significantly the robot stability and the \gls{mcot}.
Finally we demonstrate the capabilities of our \revisedtext{complete} locomotion framework (i.e. coupled planner, whole-body controller, terrain mapping and state estimation) by crossing terrains with various slopes and obstacles.
\revisedtext{All} the experimental results \revisedtext{are in} the \href{https://youtu.be/KI9x1GZWRwE}{accompanying video} or in \href{https://youtu.be/KI9x1GZWRwE}{Youtube}\footnote{\url{https://youtu.be/KI9x1GZWRwE}}.

\subsection{\revisedtext{Motion planning: decoupled vs coupled approach}}
\begin{table*}[t]
  \caption{Number of footholds, \newtext{average} walking speed and normalized \gls{mcot} for \revisedtext{different} challenging terrains \revisedtext{without changes in the elevation} for our coupled (Coup.) and decoupled (Dec.) planners.
  We normalize the \gls{mcot} with respect to the walking velocity \revisedtext{to easily compare results across different motion speeds}.
  All the results are computed from simulations.}
  \label{tab:comparison_results}
\begin{center}
\begin{tabular}{@{} lccc c ccc c ccc @{}}
\toprule
&\multicolumn{3}{c}{\# of Footholds} && \multicolumn{3}{c}{Avg. Speed [cm/s]} &&
\multicolumn{3}{c}{\newtext{ELC} / speed [s/cm]} \\
\cmidrule{2-4}\cmidrule{6-8}\cmidrule{10-12}
\emph{Terrain} 	&Coup.   &Dec.   &Ratio  &  &Coup.   &Dec.   &Ratio  &
				&Coup.   &Dec.   &Ratio  \\
\midrule
 S. Stones      &\textbf{31}     &38     &0.82	&  &\textbf{11.16}  &6.29   &1.77 
 & &13.20  &\textbf{11.43}  &1.15      \\
 Pallet 	    &\textbf{35}     &36     &0.97   &  &\textbf{9.23}   &6.92   &1.33 
 & &13.21  &\textbf{11.70}  &1.13        \\
 Stairs     	&\textbf{21}     &23     &0.91   &  &\textbf{12.79}  &11.26  &1.14 
 & &10.22  &\textbf{6.24}   &1.63        \\
 Gap			&\textbf{18}     &24     &0.75   &  &\textbf{12.76}  &9.00   &1.42
 & &9.05   &\textbf{6.84}   &1.32   \\
\bottomrule
\end{tabular}
\end{center}
\end{table*}

\begin{figure*}[tb]
	\centering
	\href{https://youtu.be/KI9x1GZWRwE#t=01m24s}{
	\includegraphics[width=1.\textwidth]{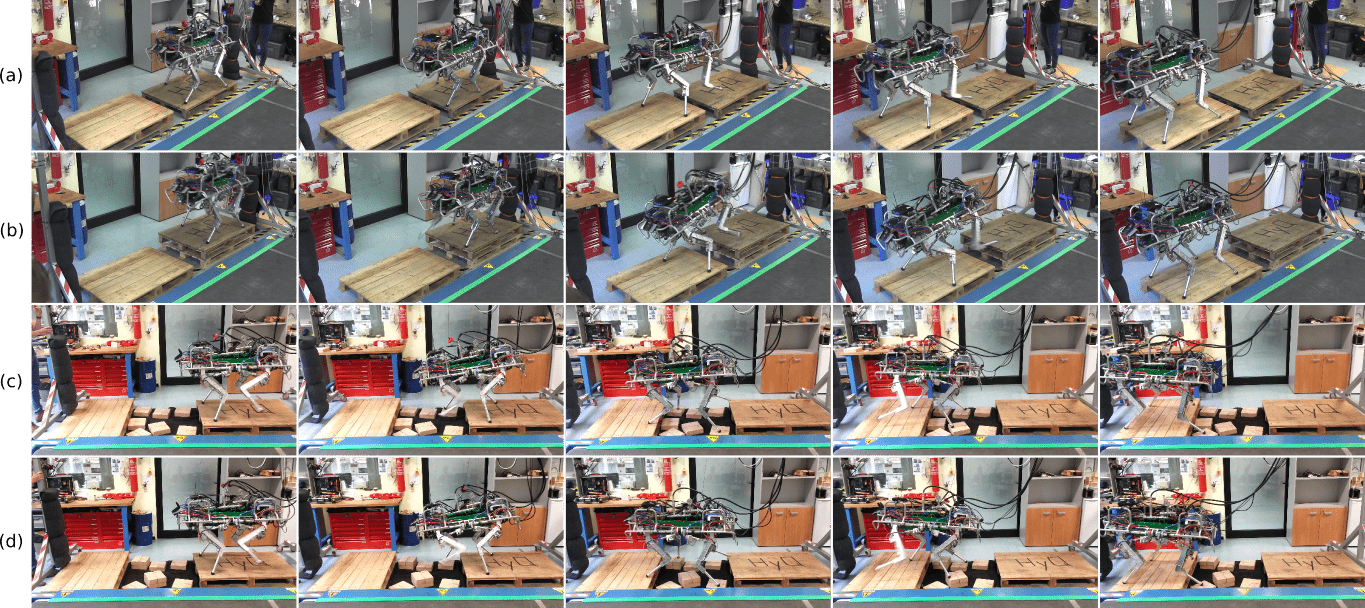}}
	\caption[Snapshots of experimental trials used to evaluate the performance of our trajectory optimization framework]{
	Snapshots of experimental trials used to evaluate the performance of \revisedtext{our trajectory optimization framework}.
	(a) crossing a gap of~\unit[25]{cm} length while climbing up~\unit[6]{cm}.
	(b) crossing a gap of~\unit[25]{cm} length while climbing down~\unit[12]{cm}.
	(c) crossing a set of 7 stepping stones.
	(d) crossing a sparse set of stepping stones with different  elevations (\unit[6]{cm}).
	To watch the video, click the figure.}
	\label{fig:rough_locomotion_snapshots}
\end{figure*}

\subsubsection{\newtext{Decoupled planner setup}}\label{sec:decoupled_setup}
\newtext{The} swing and stance duration are \revisedtext{predefined} since they cannot be optimized.
The footstep planner explores partially a set of candidate footholds using the terrain-aware heuristic function~\citep{Mastalli2015}.
\revisedtext{These} duration \revisedtext{are tuned for} every terrain \revisedtext{and}, range from 0.5 to \unit[0.7]{s} and from 0.05 to \unit[1.4]{s} for the swing and stance\footnote{In this work, with \textit{stance} phase, we refer to \revisedtext{the case when the robot has} all the feet on the ground.} phases, respectively.
\newtext{However, it \revisedtext{is} not \revisedtext{always} possible to compute a set of polynomial's coefficients \revisedtext{(\gls{com} trajectory)} that \revisedtext{satisfies} the dynamic stability for \revisedtext{some} footstep sequences.
Thus, \revisedtext{unfortunately}, step and swing duration \revisedtext{need to be \textit{hand-tuned} depending} on the footstep sequence itself.}

\subsubsection{\newtext{Coupled planner setup}}
\revisedtext{The} \textit{same} weight values \revisedtext{for the cost functions are used} for all the results presented in~\tref{tab:comparison_results} (i.e. 300, 30 and 10 for the human velocity commands, terrain, and energy, respectively).
We \revisedtext{did} not re-tune these \revisedtext{weights for a different experiment}, as it \revisedtext{was} sometimes necessary with the decoupled planner.
\revisedtext{This} shows a \revisedtext{greater} generality \revisedtext{with respect to the} decoupled planner.
\revisedtext{We add a quadratic penalization, when the terrain cost $\mathbf{T}(x,y)$ is higher then 0.8 (i.e. 80\% of its maximum value)}.

\revisedtext{For this kind of problems, it is not trivial to define a good initialization trajectory (i.e. to warm-start the optimizer).
However, since our solver uses stochastic search, this is not so critical and we decided not to do it}.
We used the same stability margin and angular acceleration (as in~\sref{subsec:attitude_results}) for the trunk attitude planner, and \revisedtext{the} horizon is $N=1$, i.e. 1 locomotion \revisedtext{cycle} or 4 steps\footnote{\revisedtext{As mentioned early, we define 6 phases which 2 of them are stance ones.}}.

\subsubsection{\newtext{Increment of the success rate}}

The foothold error is on average around \unit[2]{cm}, half \revisedtext{than in} the decoupled planner \revisedtext{case}.
\revisedtext{Note that these results are obtained} with the state estimation algorithm proposed in~\citep{Nobili2017}.
\revisedtext{The coupled planner} dramatically increases the success of the stepping stones trials to 90\%; \revisedtext{up over} 30\% with respect to the decoupled planner~\citep{Winkler2015}.
We define as \textit{success} when the robot crosses the terrain, e.g. it does not make a step in the gap, \revisedtext{and does not} reach \revisedtext{its} torque \revisedtext{and kinematic} limits.
In~\Cref{tab:comparison_results} we report the number of footholds, the \revisedtext{average} trunk speed, and the \gls{mcot} \revisedtext{for simulations made with} our coupled and decoupled planners \revisedtext{in different} challenging terrains.
\revisedtext{The coupled planner also increases} the walking velocity of least 14\% \revisedtext{and up to 63\%}, while also modulating the trunk attitude.
\revisedtext{The number of footholds is also reduced of 14\% on average.}
Jointly optimizing the motion and footholds reduces the number of \revisedtext{steps} because it considers the robot dynamics for the foothold selection.
\revisedtext{Note that the trunk speed and the success rate increased} even with terrain elevation changes (e.g. gap and stepping stones).
The \gls{mcot} is higher for our coupled planner; however, this is an effect of higher walking velocities and of the tuning of the cost function.
This is expected even if we normalized the \gls{mcot} with respect to the walking velocity.
Note that \revisedtext{as} velocity increases the kinetic energy \revisedtext{rises quadratically with a consequent affect on} the \gls{mcot}.
We also found that the tuning of the \gls{mcot} cost does not affect the stability and the foothold selection.
 
\subsubsection{\newtext{Computation time}}
An important drawback of \revisedtext{including the terrain cost-map} is that \revisedtext{it} increases substantially the computation time.
In fact for our planners, \revisedtext{this increases} from~\unit[2-3]{s} to~\unit[10-15]{min}, for more details about the computation time of the decoupled planner see~\citep{Mastalli2015}.
The main reason is that we \revisedtext{use a stochastic search which estimates the gradient} (see~\citep{Hansen2014}).
Instead, for the \revisedtext{decoupled planner}, we use a tree-search algorithm (i.e. \gls{ara}) with \revisedtext{a} heuristic function that guides the solution towards a shortest path, not the safest one, \revisedtext{which allows us to formulate the \gls{com} motion planning through a \gls{qp} program}.
\revisedtext{For more details about the footstep planner, used in the decoupled planning approach, see~\citep{Mastalli2015}.}

\subsubsection{Crossing challenging terrains}
Trunk attitude adaptation tends to overextend the legs, especially in challenging terrains, as \revisedtext{bigger} motions are required.
To avoid kinematic limits, we define a foot search region.
This ensures kinematic feasibility for terrain height difference \revisedtext{of up to \unit[12]{cm}} (coupled planner), in~\cref{fig:rough_locomotion_snapshots}\textcolor{blue_iit}{a,~b}.
Note that \revisedtext{in the decoupled case} we had to define a more conservative foot search region \revisedtext{(i.e. in the footstep planner)} than in the coupled \revisedtext{one}, making very challenging to cross gaps or stepping stones with height variations.
\revisedtext{Indeed,} crossing the terrain in~\cref{fig:rough_locomotion_snapshots}\textcolor{blue_iit}{a-d} is only possible \revisedtext{using} the coupled planner since we managed to increase the foothold region from (\unit[20]{cm}$\times$\unit[23.5]{cm}) to (\unit[34]{cm}$\times$\unit[28]{cm}).
\revisedtext{Note that the decoupled planning requires smaller foothold regions due to the fact that only considers the robot's kinematics.}

For all our optimizations, we define a stability margin of $r=$\unit[0.1]{m} (introduced in~\sref{subsec:trunk_modulation}) which is a good trade-off between modeling error and allowed trunk attitude adjustment.

\subsection{\revisedtext{Trunk attitude planning}}
\label{subsec:attitude_results}
\begin{figure}[tb]
	\centering
	\href{https://youtu.be/KI9x1GZWRwE#t=00m54s}{
	\includegraphics[width=0.48\textwidth]{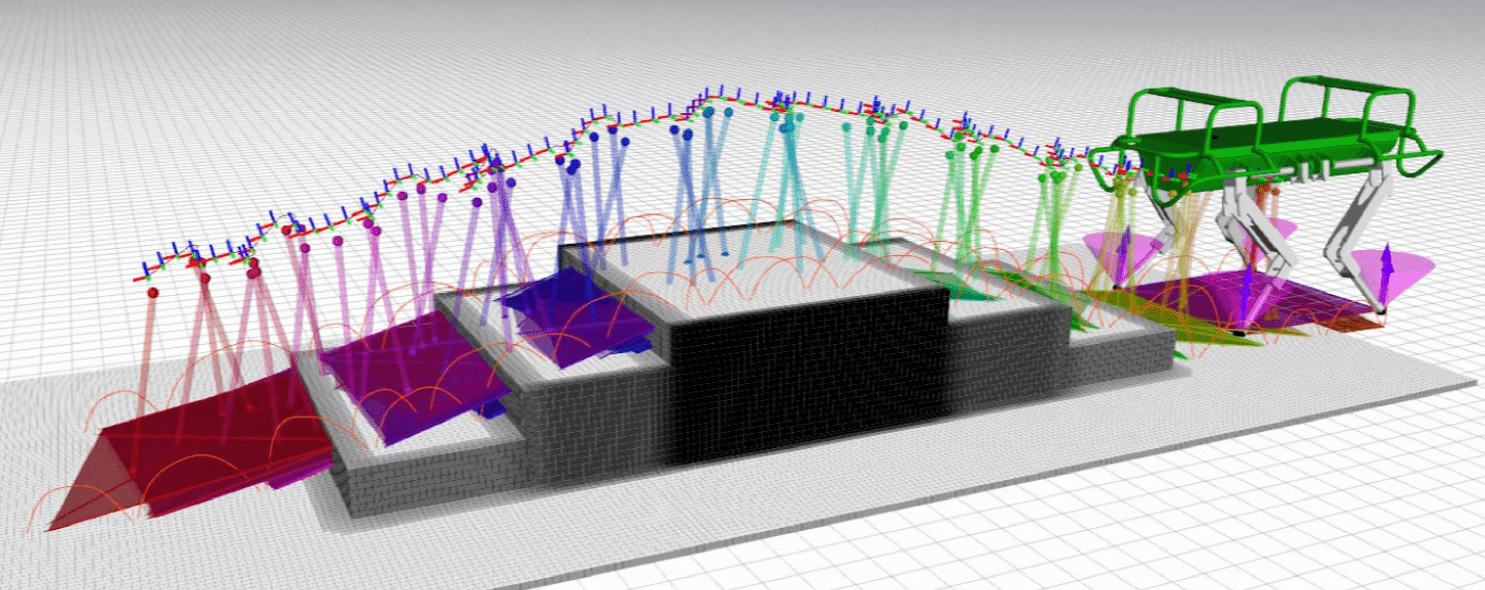}}
	\caption{
	An optimized sequence of control parameters for stair climbing.
	As in previous experiments, we use the same optimization weight values for the entire course of the motion.
	\newtext{The step heights are \unit[14]{cm}.} To watch the video, click the figure.}
	\label{fig:stairs_climbing}
\end{figure}

\begin{figure}%[htb]
	\centering
	\vspace{2mm}
	\subfloat[Dynamic walking and trunk modulation]{
	\includegraphics[width=0.46\textwidth]{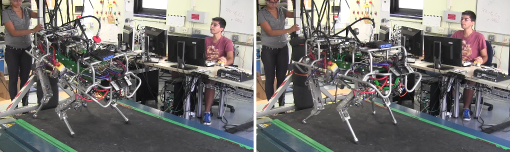}
	\label{fig:attitude_snapshots}} \\
	\subfloat[\gls{com} tracking performance]{
	\includegraphics[width=0.46\textwidth]{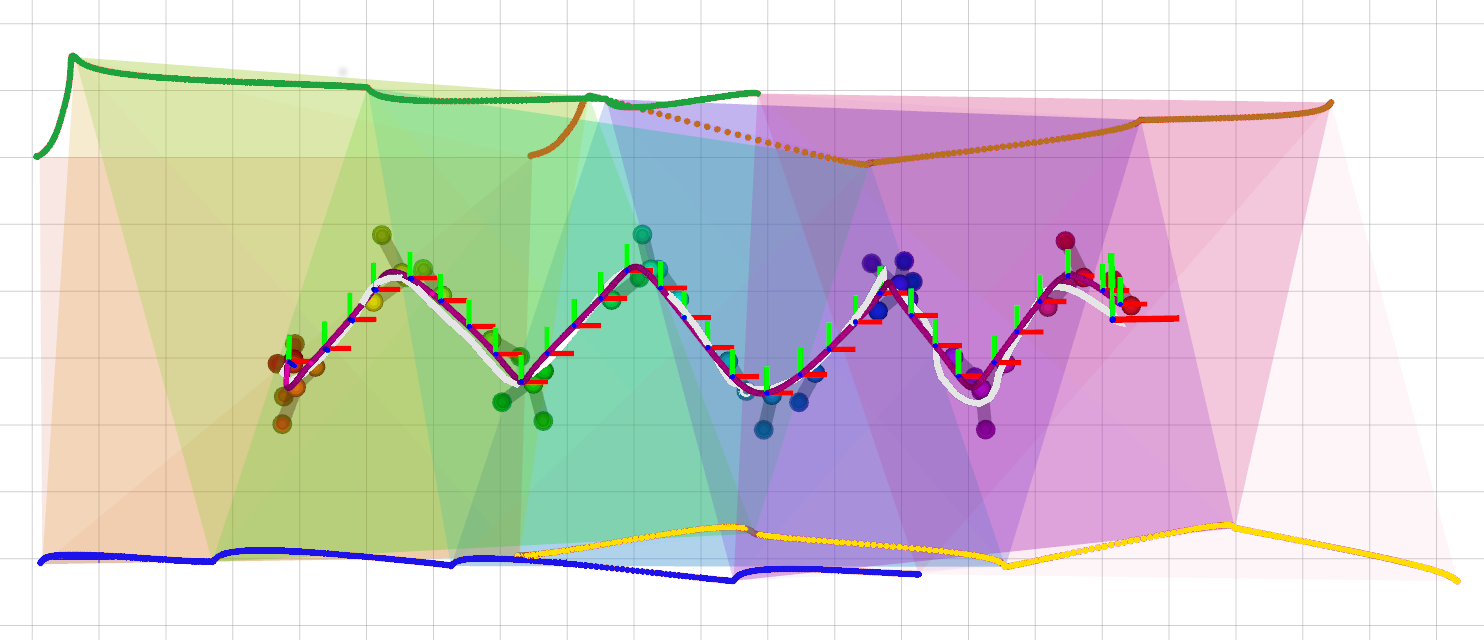}
	\label{fig:com_tracking_top}} \\
	\subfloat[Trunk attitude modulation]{
	\includegraphics[width=0.46\textwidth]{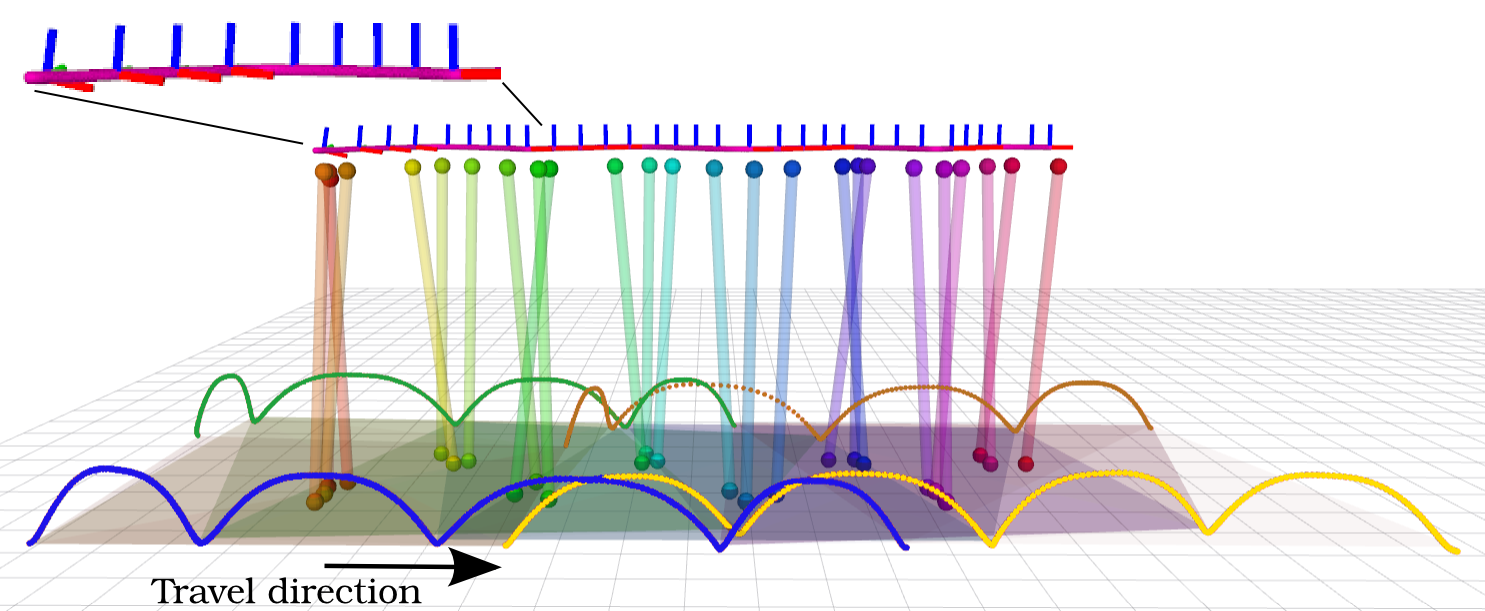}
	\label{fig:com_tracking_lateral}}
	\caption[Dynamic attitude modulation]{
	(a) Dynamic attitude modulation \revisedtext{on the \gls{hyq} robot}.
	The initial trunk attitude is 0.17 and 0.22 radians in roll and pitch, respectively.
	(b) Body tracking when walking and dynamically modulating the trunk attitude.
	The planned \gls{com} (magenta) and the executed trajectory (white) are shown together with the sequence of support polygons, \gls{cop} and \gls{com} positions.
	Note that each phase is identified with a specific color.
	(c) A lateral view of the same motion shows the attitude correction (sequence of frames), and the cart-table displacement.
	\newtext{We} use the RGB color convention for drawing the different frames. In (b)-(c) the brown, yellow, green and blue trajectories represent the \gls{lf}, \gls{rf}, \gls{lh} and \gls{rh} foot trajectories, respectively.}
\end{figure}

\revisedtext{The \textit{cart-table} model neglects the angular dynamics and therefore cannot be used to control the robot's attitude.}
However, \revisedtext{with a flywheel extension as proposed for our attitude planning approach, we could generate stable motions while changing the robot attitude (e.g. stair climbing as in~\fref{fig:stairs_climbing}).}
\revisedtext{In this section, we} showcase the automatic trunk attitude modulation during a dynamic walk \revisedtext{on the \gls{hyq} robot (\cref{fig:attitude_snapshots})}.

To \revisedtext{experimentally validate} the attitude modulation method, we plan a \textit{fast}\footnote{Compared to the common walking-gait velocities of \gls{hyq}.} dynamic walk with a trunk velocity of \unit[18]{cm/s}, \revisedtext{and an} initial trunk attitude of 0.17 and 0.22 radians in roll and pitch, respectively.
We do not use the terrain \revisedtext{cost-map} \revisedtext{to generate} the corresponding footholds, thus the resulting feet locations come from the dynamics of walking itself, while maximizing the stability of the gait.
We compute the maximum allowed angular acceleration given the trunk inertia matrix of \gls{hyq}, from~\eref{eq:max_ang_acc}, which results in \unit[0.11]{rad/s$^2$} as the maximum diagonal element.
The trunk attitude planner uses this maximum allowed acceleration to align the trunk and support plane through cubic polynomial splines (~\sref{subsec:trunk_modulation}).

\begin{figure}[h]
	\centering
	\includegraphics[width=0.48\textwidth]{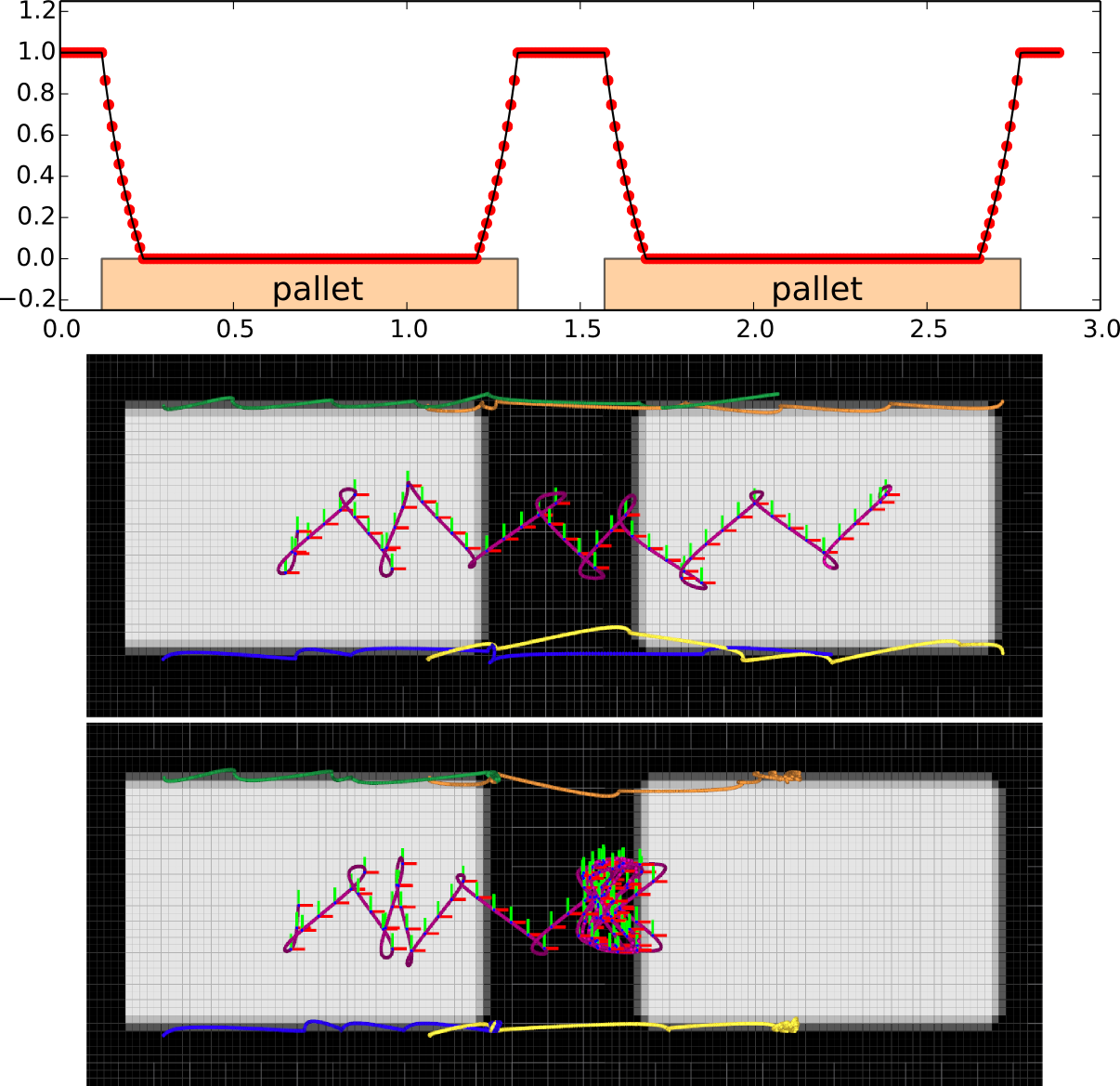}
	\caption[The effect of the terrain cost-map]{
	The effect of changing terrain weight values when crossing a gap of \unit[25]{cm}.
	The \revisedtext{cost-map} is computed only using the height deviation	feature (\textit{top}); the red points represent the discretization of the continuous cost function (\unit[1]{cm}).
	\revisedtext{The cost values are represented using gray scale, where white and black are the minimum and maximum cost values, respectively.
	A higher value in the terrain weight describes a higher risk for foothold locations near the borders of the gap.
	An appropriate weight allows the robot to cross the gap (\textit{middle}).}
	In contrast, an	increment of 200\% in the weight penalizes excessively footholds close to the gap and as result the robot cannot cross the gap as kinematic
	limits are exceeded (\textit{bottom}).}
	\label{fig:terrain_costmap_effect}
\end{figure}

The resulting behavior shows the \gls{hyq} robot successfully walking while changing its trunk roll and pitch angles.
The trunk attitude planner adjusts the roll and pitch angles given the estimated support region at each phase.
\cref{fig:com_tracking_top} shows the \gls{com} tracking performance for \revisedtext{an} initial trunk attitude of \unit[0.17]{rad} and \unit[0.22]{rad} in roll and pitch,
respectively.
\cref{fig:com_tracking_lateral} shows that the entire attitude modulation is accomplished in the first 6 phases (i.e. one \revisedtext{locomotion} cycle or four steps \revisedtext{with two support phases}).
Because our attitude planner \newtext{keeps the \gls{cmp} inside the support region}, the \gls{hyq} robot successfully \revisedtext{crosses} terrains with \revisedtext{different heights} as shown in~\fref{fig:rough_locomotion_snapshots}\textcolor{blue_iit}{a-d}.
The stability margin is the same for all the experiments in this paper ($r=0.1$ m).

\subsection{\revisedtext{Motion planning and terrain mapping}}
Different weighting choices on the terrain cost-map produce \revisedtext{different} behaviors, \revisedtext{as described} by~\eref{eq:terrain_weight}.
For simplicity, we analyze the effect of these weights \revisedtext{in} gap crossing.
\revisedtext{We observe two different plans which are only influenced by the terrain weight in the cost function (\fref{fig:terrain_costmap_effect}).}
Strongly penalizing the terrain \revisedtext{cost-map} results in the robot not being able to cross the gap due to its kinematic limits (\fref{fig:terrain_costmap_effect}\textcolor{blue_iit}{(\textit{bottom})}).
By reducing the terrain weight, \revisedtext{we observe} that the coupled planner \revisedtext{selects} footholds closer to the gap border, which allows the robot to cross the \revisedtext{space} (\fref{fig:terrain_costmap_effect}\textcolor{blue_iit}{(\textit{top})}).
The terrain weight mainly influences the foothold selection, and does not influence the stability or the \gls{mcot}.
\revisedtext{Finally, we observed that the computational cost is not affected by the terrain geometry.}
\begin{figure*}[tb]
	\centering
	\href{https://youtu.be/KI9x1GZWRwE#t=01m59s}{
	\includegraphics[width=1.\textwidth]{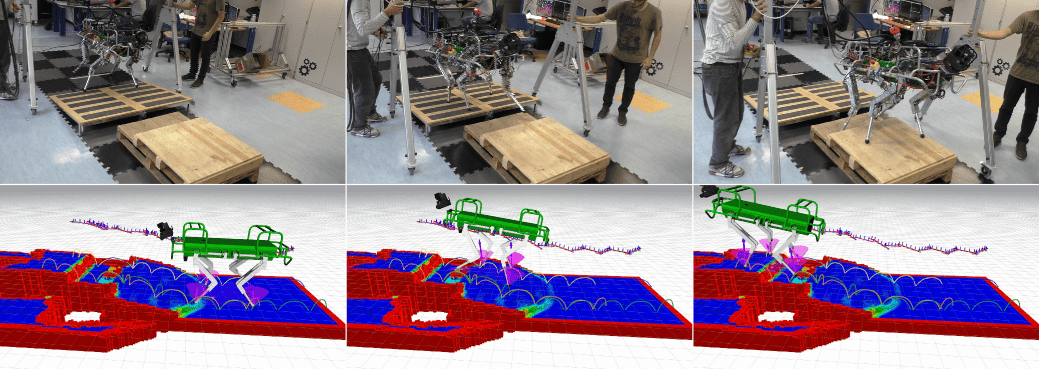}}
	\caption{
	Crossing a terrain that combines elements of the previous cases; first a ramp of $10$ degree, then a gap of \unit[15]{cm} and finally a step with \unit[15]{cm} height \revisedtext{change}.
	Execution of the planned motion with the \gls{hyq} robot (top).
	Visualization of the terrain cost-map, friction cone and \gls{grfs} (bottom).
	The color for the friction cone and \gls{grfs} are magenta and purple, respectively.
	To watch the video, click the figure.}
	\label{fig:sloppy_snapshots}
\end{figure*}

\subsection{\revisedtext{Whole-body control, state estimation and terrain mapping}}
\revisedtext{The} whole-body controller \revisedtext{successfully tracks} the planned motion without violating friction, torques or kinematics constraints \fref{fig:sloppy_snapshots}\textcolor{blue_iit}{(\textit{bottom}).}
\revisedtext{A key aspect is that our controller follows the desired wrenches computed from the motion plan, giving priority to the above constraints.
This is important because our coupled planner does not consider the non-coplanar contact condition and friction cone (since the used cart-table model \revisedtext{that} neglects them).
With this approach, the robot can (1) climb in simulation ramps up to $20$ degrees in similar friction conditions to real experiments ($\mu=0.7$) and (2) handle unpredictable contact interactions as shown in~\fref{fig:stairs_climbing}}.

\begin{figure}[t]
	\centering
	\includegraphics[width=0.98\columnwidth]{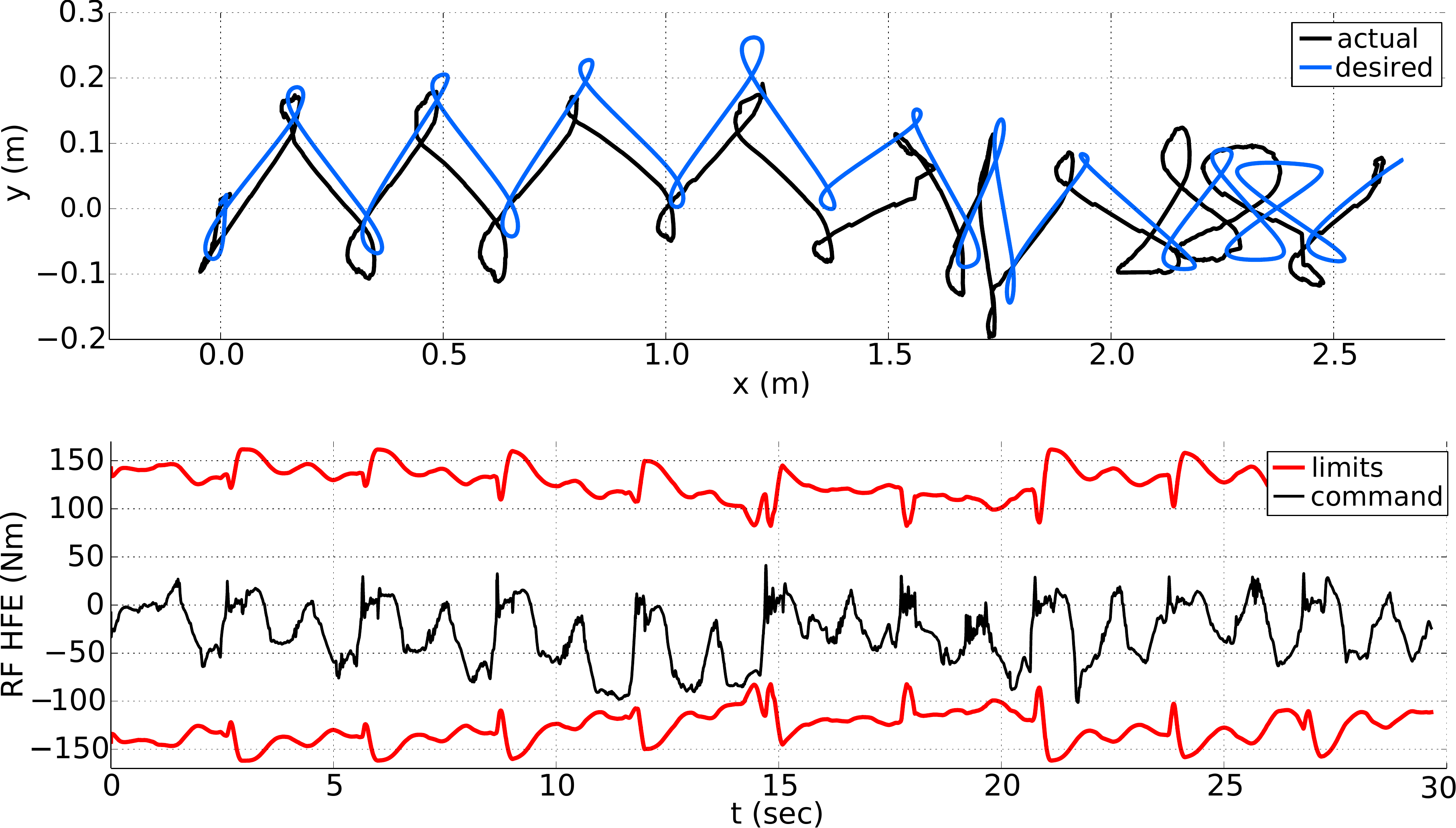}
	\caption{
	\gls{hyq} crossing a terrain that combines elements of all previous cases.
	(\textit{Top}): \gls{com} tracking performance, desired (blue) and executed (black) motions.
	(\textit{Bottom}): applied torque command along the	course of the motion. At $t=$\unit[14]{sec}, the planned motion produced a movement	that reached the torque limits; however, the controller applies a torque command inside the robot's limits.
	In fact, the tracking error increases at approximately $x=$\unit[1.25]{m}, and is reduced in the next steps.}
	\label{fig:sloppy_plots}
\end{figure}

The terrain surface normals are computed online from vision.
The \revisedtext{friction} coefficient used in \revisedtext{these} trials (i.e. simulation and experiments) is $0.7$, which is a conservative \revisedtext{estimate} of the real contact conditions.
\revisedtext{\fref{fig:sloppy_plots} shows the tracking performance against errors in the state estimation and terrain mapping.
The tracking error is mainly due to low-frequency corrections of the estimated pose.}

%------------------------------ DISCUSSION
% %%%%%%%%%%%%%%%%%%%%%%%%%%%%%%%%%%%%%%%%%%%%%%%%%%%%%%%%%%%%%%%%%%%%%%%%%%%%%%%
% 2345678901234567890123456789012345678901234567890123456789012345678901234567890
% 1         2         3         4         5         6         7         8
% %%%%%%%%%%%%%%%%%%%%%%%%%%%%%%%%%%%%%%%%%%%%%%%%%%%%%%%%%%%%%%%%%%%%%%%%%%%%%%%
% ******************************************************************************
% Discussion
% ******************************************************************************
\section{Discussion}\label{sec:discussion}
\subsection{\revisedtext{Motion planning: decoupled vs coupled approach}}
Coupled motion and foothold planning \revisedtext{include} dynamics \revisedtext{in} foothold selection.
\revisedtext{This is critical to both increase the range of possible foothold locations and to adjust the step duration.}
Both \revisedtext{of these parameters} allow the robot to cross a \revisedtext{wider} range of terrains.
We \revisedtext{noted} that the coupled planner handles \revisedtext{different} terrain \revisedtext{heights} more easily because of the joint optimization process.
Crossing gaps with various elevations exposed the limitation of decoupled methods, since \revisedtext{this is more prune to hit the kinematic limits} (see~\cref{fig:rough_locomotion_snapshots}\textcolor{blue_iit}{a}).
However, an important drawback of coupled foothold and motion planning is the \revisedtext{increase in} computation time compared \revisedtext{to} decoupled planning.
It is possible to reduce the computation time by describing the foothold \revisedtext{using} integer variables~\citep{Deits2014,Aceituno2017b}, but this would \newtext{limit} the \newtext{number of \revisedtext{feasible convex} regions}.
Instead the coupled planning uses a terrain model that considers a broader range of challenging environments \revisedtext{}{because of the ``continuous'' cost-map}.
In any case, the computation time remains longer for coupled planning as we presented in~\citep{Aceituno2017b}.

\newtext{Optimizing the step timing has not shown a clear benefit in our experimental results.
\revisedtext{We argue that step timing is important to find feasible solutions when there is a small friction coefficient or the risk of reaching torque limits, i.e. a slower motion is needed to satisfy both constraints.
However, the cart-table model does not consider these constraints, which in practice makes the time optimization not useful for reduced dynamics.}}

\begin{figure}[tb]
	\centering
	\includegraphics[width=0.48\textwidth]{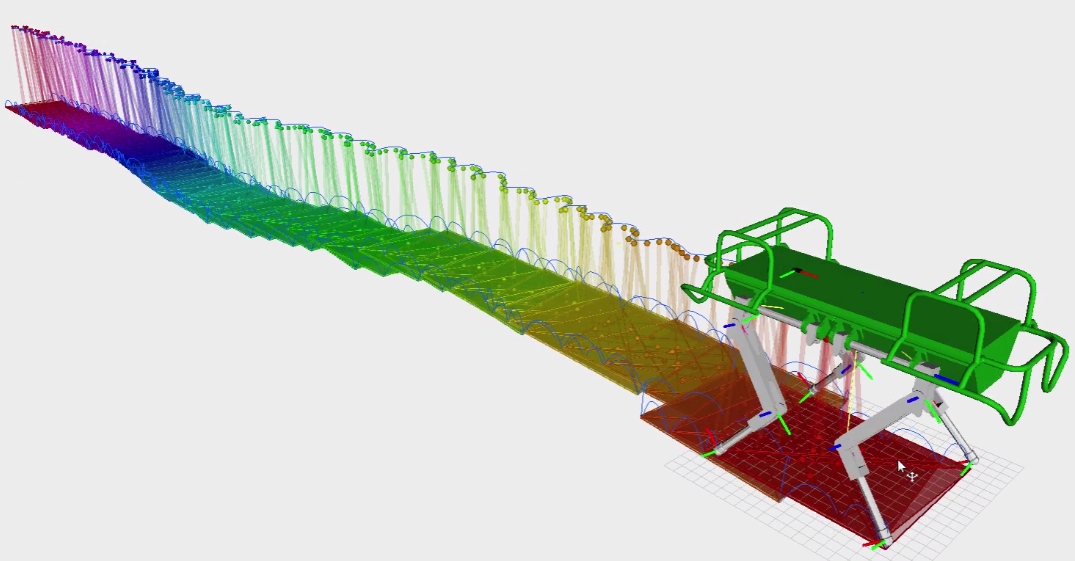}
	\caption{
    \newtext{Optimizing 208 steps from 52 trajectory optimization problems.
    In each optimization problem the step timing, foothold locations and \gls{com} trajectory for 4 steps in advance with different velocity commands \revisedtext{are computed}.
    The color describes different step phases of the planned motion.}}
	\label{fig:208_steps}
\end{figure}

\newtext{We have tested, in simulation, our \revisedtext{locomotion framework} up to 208 steps \revisedtext{on} flat terrain (\fref{fig:208_steps}) \revisedtext{and up to 50 steps in non-flat terrain (\fref{fig:stairs_climbing})}.
The \revisedtext{modeling} errors on the cart-table with flywheel approximation are easily handled by the whole-body controller.
In addition, the swing trajectory are expressed in the base frame, so errors in the state estimation affect little the stability.
However, unexpected events can compromise the stability (e.g. unstable footsteps, moving obstacles, state estimation errors as a result of slippage, etc) \revisedtext{and re-planning might be needed}.
According to our experience, it is recommended to optimize at least one cycle of locomotion since we do not know the \gls{com} travel direction and velocity in an individual step.
For all the experiments, we plan 4 steps ahead and it was not needed a longer horizon.
We planned 6 and 8 steps ahead without any significant improvement in the motion.
\revisedtext{To compute the whole motion, we solve different trajectory optimization problems in receding fashion.}
}

% \newtext{
% 	\todo{Im confused the next paragraph talks about bellicoso work? is a bit unconnected, then you say with the douepled approach was impossible to cross gaps ramps stepping stones, that is not consistent with table I}
% 	In~\citep{Bellicoso2017}, the \gls{com} trajectory is represented with polynomial splines and is solved through a \gls{qp} optimization \revisedtext{similar} to our decoupled planning approach~\citep{Winkler2015}.
% \revisedtext{The} main \revisedtext{difference} lies in the foothold selection that in the decoupled approach is computed heuristically given the desired robot velocity.
% This is very restrictive, and \revisedtext{makes crossing} stepping stones, gaps, stairs, ramps, \revisedtext{etc. impossible}, since it is \revisedtext{necessary} to carefully plan footholds given the terrain map and to account for the robot stability against changes in the body attitude.
% \revisedtext{Associated with this is a need} to hand tune the footholds and step timing to properly fit the polynomial (see~\sref{sec:decoupled_setup}).
% We also see an \revisedtext{increase} in the number of footsteps, as reported in~\tref{tab:comparison_results}.}

\subsection{\revisedtext{Trunk attitude planning}}
The cart-table model estimates the \gls{cop} position, yet it neglects the angular components of the body motion \revisedtext{that} can lead to \revisedtext{inaccuracies in the \gls{cop} estimation}.
This can affect the stability \revisedtext{particular when there is a change in height e.g. climbing/descending gaps or stairs}, crossing uneven stepping stones, etc.
To systematically address these effects without affecting the stability, a relationship \revisedtext{was obtained} between the torques \revisedtext{applied} to the \gls{com} and the displacement of the \gls{cop}.
Later, we connected the stability margin by assuming a time-invariant inertial tensor approximation of the inertia matrix.
Experimental results with the \gls{hyq} robot validated this method for challenging terrain locomotion.
\revisedtext{The method developed in this paper} can be applied to other legged systems, such as humanoids.

\revisedtext{Our attitude planner does not aim to control angular momentum, instead we propose 1) to use a heuristic for trunk orientation and 2) to guarantee the robot stability under mild assumptions.
To handle the \textit{zero-dynamics} instabilities (described in~\citep{Nava2016}), our whole-body controller tracks the desired robot orientation computed by the trunk attitude planner.
However, we argue that a more effective robot attitude planner will require to consider the limb dynamics (full dynamics) and to account for future events (planning).
It is clearly crystallized in the \textit{cat-falling} motion, when the momenta conservation defines a nonholonomic constraint on the angular momentum (for more details see~\citep{Wieber2006}).
This is the reason why recent works have been focused on efficient full-body optimization (e.g.~\cite{Budhiraja2018,Herzog2016,Mastalli2020}.
}

\subsection{The effect of the terrain cost-map}
Considering the terrain topology increases the complexity of the trajectory optimization problem.
Moreover, optimizing the step duration introduces many local minima in the problem landscape.
\revisedtext{To address} these issues, a low-dimensional parameterized model \revisedtext{is used} which allows us to \revisedtext{use stochastic-based search}.
\revisedtext{Note that stochastic-based search becomes quickly intractable when the problem dimension increases.}
Even \revisedtext{though} our problem is non-convex, we \revisedtext{reduce} the number of required footholds by an average of 13.75\% compared to our convex decoupled planner
(\tref{tab:comparison_results}).
\revisedtext{The terrain cost-map increases the robustness of the planned motion.}
\revisedtext{Indeed,} the selected footsteps are far from risky regions\revisedtext{, and this is very important to increase robustness because} tracking errors always produce variation on the executed footstep.

\subsection{Considering terrain with slopes}
Higher walking speed increases the probability of foot-slippage.
When one or \revisedtext{more} of the feet slip backwards, or when a foot is only slightly loaded, \revisedtext{might result in a poor tracking}.
Both events are more likely to happen in a terrain with different elevations due to errors in the state estimation or noise in the \newtext{exteroceptive} sensors.
Including friction-cone and \revisedtext{foot unloading / loading} constraints\revisedtext{, in the whole-body optimization, has \revisedtext{been} shown to help mitigate the poor tracking}.
We demonstrated experimentally that is possible to navigate a wide range of terrain slopes without considering the friction cone stability in the planning level \revisedtext{(only at the controller level)}.

\subsection{Terrain mapping and state estimation}
Estimating the state of the robot with a level of accuracy suitable for planned motions \revisedtext{is} a challenging task.
Reliable state estimation is crucial, as accurate foot placement directly depends on the robot's base pose estimate.
The estimate is also used to compute the \revisedtext{desired} torque commands through a virtual model.
The major sources of error for inertial-legged state estimation are IMU gyro bias and foot slippage.
These produce a pose estimate drift, which \revisedtext{cannot be completely eliminated by the contact state estimate~\citep{Camurri2017} (or with proprioceptive sensing).}
\revisedtext{Pose} drift particularly affects the \revisedtext{desired} torques computed from the \revisedtext{whole-body optimization}.
To eliminate it, we fused high frequency (\unit[1]{kHz}) proprioceptive sources (inertial and leg odometry) with low frequency exteroceptive updates (\unit[0.5]{Hz} for LiDAR \revisedtext{registration}, \unit[10]{Hz} for \revisedtext{optical flow}) in a combined Extended Kalman Filter~\citep{Nobili2017}.
We noticed that the drift accumulated, in between the high frequency proprioceptive updates and the low frequency exteroceptive updates, affected the \revisedtext{experimental performance}.
In practice, to cope with this problem, we reduced the compliance of our whole-body controller (by increasing the proportional gains).

%------------------------------ CONCLUSION
\section{Conclusion}\label{sec:conclusion}
In this paper, we \revisedtext{presented a new} framework for dynamic whole-body locomotion on challenging terrain.
We \revisedtext{extended} our previous planning approach from~\citep{Mastalli2017} \revisedtext{by modeling} \newtext{the terrain through log-barrier functions in the numerical optimization.
In addition, we \revisedtext{proposed a} novel robot attitude planning \revisedtext{algorithm}.
\revisedtext{Using} this, we \revisedtext{could} optimize both the \gls{com} motion and footholds in the horizontal frame, and \revisedtext{allow the robot to adapt its trunk orientation}.
\revisedtext{We demonstrated in experimental trials and simulations that the assumptions on the attitude planner avoided instability under significant terrain elevation changes (up to 12 $cm$).}
We compared \revisedtext{coupled and decoupled planning} and highlighted the advantages and disadvantages of \revisedtext{them}.}
In our test-case planners, we \revisedtext{used the same} method for quantifying the terrain difficulty (i.e. terrain cost-map).
We showed that reduced models for motion planning (such as cart-table \newtext{with flywheel}) \revisedtext{together with whole-body control} are still \textit{suitable} for a wide range of challenging scenarios.
\revisedtext{We used the full dynamic model only in our real-time whole-body controller to avoid slippage, and hitting torque and kinematic limits.
The online terrain mapping allowed our controller to avoid slippage on the trialled terrain surfaces.}
\revisedtext{We presented results, validated by experimental trials and comparative evaluations, in a series of terrains of progressively increasing complexity}.

\bibliography{reference}
\printglossaries
\vspace{-1cm}
\begin{IEEEbiography}[{\includegraphics[width=1in,height=1.25in,clip,keepaspectratio]{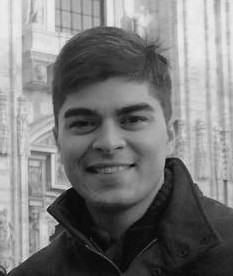}}]%
{Carlos Mastalli} received the Ph.D. degree on ``Planning and Execution of Dynamic Whole-Body Locomotion on Challenging Terrain'' from Istituto Italiano di Tecnologia, Genoa, Italy, in April 2017.

He is currently a Research Associate in the University of Edinburgh with Alan Turing fellowship.
His research combines the formalism of model-base approach with the exploration of vast robot’s data for robot locomotion.
From 2017 to 2019, he was a postdoc in the Gepetto Team at LAAS-CNRS.
Previously, he completed his Ph.D. on ``Planning and Execution of Dynamic Whole-Body Locomotion on Challenging Terrain'' in April 2017 at Istituto Italiano di Tecnologia.
He is also improved significantly the locomotion framework of the HyQ robot.
His has contributions in optimal control, motion planning, whole-body control and machine learning for legged locomotion.
\end{IEEEbiography}
\vspace{-1cm}
\begin{IEEEbiography}[{\includegraphics[width=1in,height=1.25in,clip,keepaspectratio]{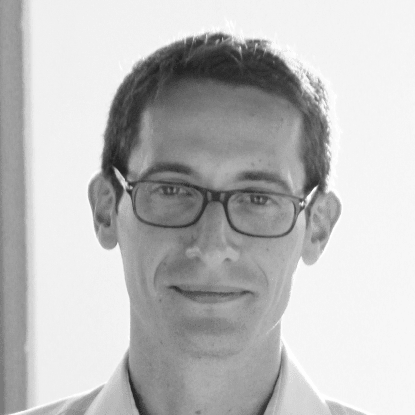}}]%
{Ioannis Havoutis} received the M.Sc. in Artificial Intelligence and Ph.D. in Informatics degrees from the University of Edinburgh, Edinburgh, U.K.

He is a Lecturer in Robotics at the University of Oxford.
He is part of the Oxford Robotics Institute and a co-lead of the Dynamic Robot Systems group.
His focus is on approaches for dynamic whole-body motion planning and control for legged robots in challenging domains.
From 2015 to 2017, he was a postdoc at the Robot Learning and Interaction Group, at the Idiap Research Institute.
Previously, from 2011 to 2015, he was a senior postdoc at the Dynamic Legged System lab the Istituto Italiano di Tecnologia.
He holds a Ph.D. and M.Sc. from the University of Edinburgh.
\end{IEEEbiography}
\vspace{-1cm}
\begin{IEEEbiography}[{\includegraphics[width=1in,height=1.25in,clip,keepaspectratio]{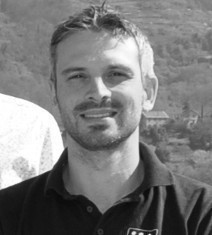}}]%
{Michele Focchi} received the B.Sc. and M.Sc. degrees in control system engineering from Politecnico di Milano, Milan, Italy, in 2004 and 2007, respectively,
and the Ph.D. degree in robotics in 2013.

He is currently a Researcher at the Advanced Robotics department, Istituto Italiano di Tecnologia.
He received both the B.Sc. and the M.Sc. in Control System Engineering from Politecnico di Milano in 2004 and 2007, respectively.
In 2009 he started to develop a novel concept of air-pressure driven micro-turbine for power generation in which he obtained an international patent.
In 2013, he got a Ph.D. degree in Robotics, where he developed low-level controllers for the Hydraulically Actuated Quadruped (HyQ) robot.
Currently his research interests are dynamic planning, optimization, locomotion in unstructured/cluttered environments, stair climbing, model identification and
whole-body control.
\end{IEEEbiography}
\vspace{-1cm}
\begin{IEEEbiography}[{\includegraphics[width=1in,height=1.25in,clip,keepaspectratio]{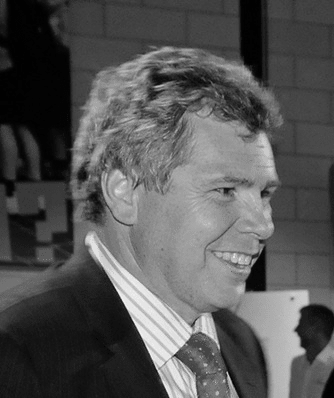}}]%
{Darwin G. Caldwell} (Senior Member, IEEE) received the B.Sc. and Ph.D. degrees in robotics from University of Hull, Hull, U.K., in 1986 and 1990, respectively, and the M.Sc. degree in management from University of Salford, Salford, U.K., in 1996.

He is a founding Director at the Istituto Italiano di Tecnologia in Genoa, Italy, and a Honorary Professor at the Universities of Sheffield, Manchester, Bangor, Kings College, London and Tianjin University China.
His research interests include innovative actuators, humanoid and quadrupedal robotics and locomotion (iCub, HyQ and COMAN), haptic feedback, force augmentation exoskeletons, dexterous manipulators, biomimetic systems, rehabilitation and surgical robotics, telepresence and teleoperation procedures.
He is the author or co-author of over 450 academic papers, and 17 patents and has received awards and nominations from several international journals and conferences.
\end{IEEEbiography}
\vspace{-1cm}
\begin{IEEEbiography}[{\includegraphics[width=1in,height=1.25in,clip,keepaspectratio]{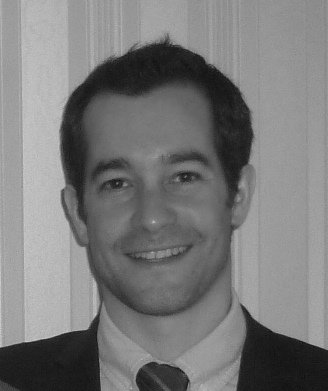}}]%
{Claudio Semini} received a M.Sc. degree in Electrical Engineering and Information Technology from ETH Zurich, Switzerland, in 2005.

He is the Head of the Dynamic Legged Systems (DLS) laboratory at Istituto Italiano di Tecnologia (IIT).
From 2004 to 2006, he first visited the Hirose Laboratory at Tokyo Tech, and later the Toshiba R\&D Center, Japan.
During his doctorate from 2007 to 2010 at the IIT, he developed the hydraulic quadruped robot HyQ and worked on its control.
After a postdoc with the same department, in 2012, he became the Head of the DLS lab. His research interests include the construction and control of versatile, hydraulic legged robots for real-world environments.
\end{IEEEbiography}\vfill

\end{document}